\documentclass[letterpaper]{article} 
\usepackage{aaai2026}  
\usepackage{times}  
\usepackage{helvet}  
\usepackage{courier}  
\usepackage[hyphens]{url}  
\usepackage{graphicx} 
\usepackage{natbib}  
\usepackage{caption} 
\usepackage{algorithm}
\usepackage{algorithmic}

\usepackage{amsmath}
\usepackage{amssymb}
\usepackage{multirow}
\usepackage{threeparttable}
\usepackage{booktabs}

\usepackage{amsthm}

\usepackage[font=small]{subcaption} 

\usepackage{xcolor}

\usepackage{newfloat}
\usepackage{listings}
\DeclareCaptionStyle{ruled}{labelfont=normalfont,labelsep=colon,strut=off} 
\floatstyle{ruled}
\newfloat{listing}{tb}{lst}{}
\floatname{listing}{Listing}
\title{FedRecon: Missing Modality Reconstruction in Heterogeneous Distributed Environments}
\author {
    Junming Liu\textsuperscript{\rm 1,2},
    Yanting Gao\textsuperscript{\rm 1},
    Yifei Sun\textsuperscript{\rm 2},
    Yufei Jin\textsuperscript{\rm 1},
    Yirong Chen\textsuperscript{\rm 2},
    Ding Wang\textsuperscript{\rm 2},
    Guosun Zeng\textsuperscript{\rm 1}\thanks{Corresponding author.},
}
\affiliations {
    \textsuperscript{\rm 1}Tongji University\\
    \textsuperscript{\rm 2}Shanghai Artificial Intelligence Laboratory\\
    liu\_junming6917@tongji.edu.cn, gszeng@tongji.edu.cn
}

\usepackage{bibentry}

\begin{document}

\maketitle

\begin{figure*}[t]
\centering
  \includegraphics[width=\textwidth]{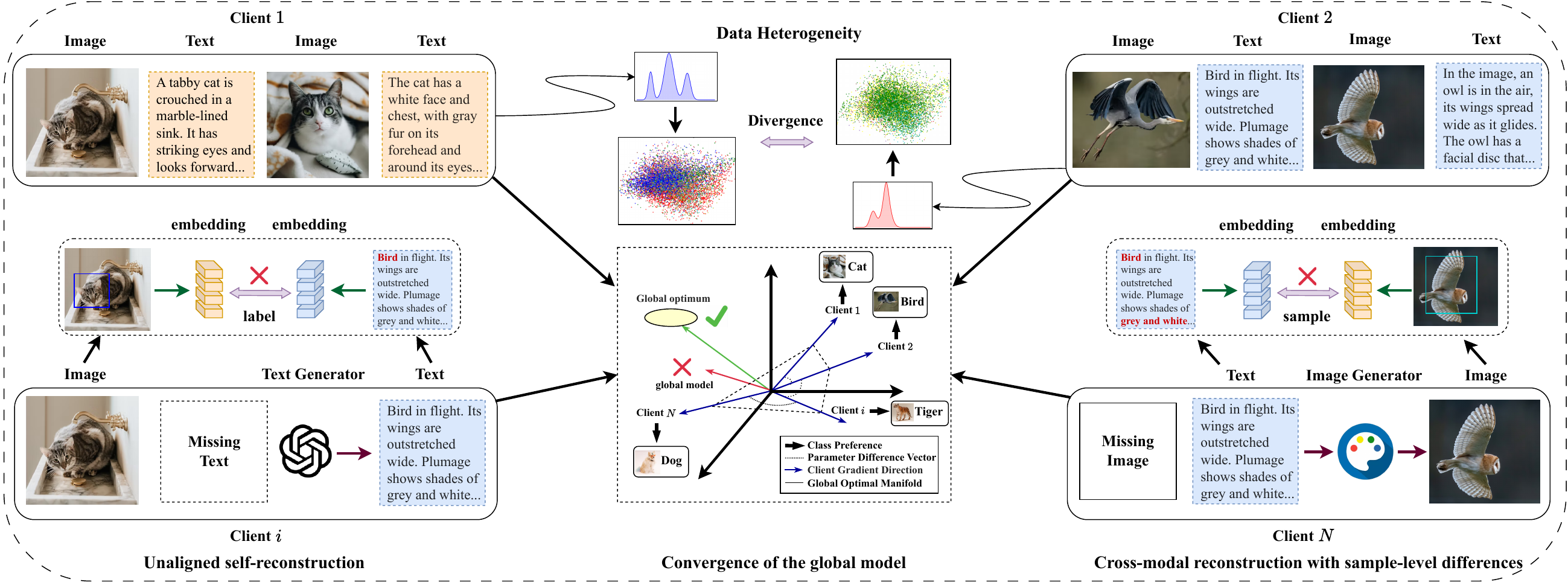}
  \caption{Impact of modality and data heterogeneity on the convergence of the global model in MFL. Client 1 and 2 have complete modalities but suffer from data heterogeneity due to divergent local distributions. In contrast, client $i$ and client $N$ are missing modalities and perform unimodal self-reconstruction and cross-modal reconstruction respectively.}
  \label{fig:problems}
\end{figure*}

\begin{abstract}
Multimodal data are often incomplete and exhibit Non-Independent and Identically Distributed (Non-IID) characteristics in real-world scenarios.
These inherent limitations lead to both modality heterogeneity through partial modality absence and data heterogeneity from distribution divergence, creating fundamental challenges for effective federated learning (FL).
To address these coupled challenges, we propose FedRecon, the first method targeting simultaneous missing modality reconstruction and Non-IID adaptation in multimodal FL.
Our approach first employs a lightweight Multimodal Variational Autoencoder (MVAE) to reconstruct missing modalities while preserving cross-modal consistency.
Distinct from conventional imputation methods, we achieve sample-level alignment through a novel distribution mapping mechanism that guarantees both data consistency and completeness.
Additionally, we introduce a strategy employing global generator freezing to prevent catastrophic forgetting, which in turn mitigates Non-IID fluctuations.
Extensive evaluations on multimodal datasets demonstrate FedRecon's superior performance in modality reconstruction under Non-IID conditions, surpassing state-of-the-art methods.
The code will be released upon paper acceptance.
\end{abstract}

\section{Introduction}

Federated Learning (FL) has gained significant attention as a promising framework for decentralized machine learning, ensuring privacy protection in distributed environments \cite{Li_2020_Review, Li_2020_InvisibleFL, Qin_2023_Privacy, Xiong_2024_Efficient, Liu_2025_Mosaic}. While existing FL methods demonstrate strong performance in unimodal settings \cite{Chen_2025_Can, Yu_2024_Addressing}, their inability to capture cross-modal interactions fundamentally limits real-world applicability \cite{Huang_2021_What, Huang_2024_On, Lin_2023_Federated}.
To bridge this gap, Multimodal Federated Learning (MFL) has emerged as a critical solution by enabling complementary information fusion across modalities \cite{Zhao_2022_Multimodal, Che_2023_Survey}, with successful deployments in autonomous driving \cite{Cui_2019_Multimodal}, video action capture \cite{Ofli_2013_Berkeley}, and multimedia sentiment analysis \cite{Kaur_2019_Multimodal}.
By explicitly modeling inter-modal dependencies, MFL achieves holistic understanding of complex phenomena \cite{Li_2024_Cross, Chen_2024_FedMBridge}, enabling precise real-world inference under data scarcity.

The majority of existing MFL approaches presume full modality availability across clients \cite{Huang_2024_Multimodal, Gao_2025_Fusion}. However, this assumption is often misaligned with real-world conditions, where data availability is frequently restricted.
For instance, certain clients may possess limited resources or only support unimodal data acquisition, whereas others may have access to more advanced multimodal setups \cite{Poria_2017_Review, Liu_2025_VaLiK}.
Furthermore, the quality of the data can be severely impacted by fluctuations during the collection process, leading to inconsistencies that undermine the reliability of the captured information \cite{Hassan_2021_Multimodal}.
In this study, we formally consider a MFL system with $K$ clients and up to $|M|$ modalities. All clients and the central server share a common model architecture, but each client $i$ has access to a subset $M_i \subseteq M$ of the available modalities. This results in a scenario where $M_i \neq M_{i'}$ for some clients $i, i' \in [K]$.
Consequently, the system introduces data heterogeneity across clients and disrupts inter-modal connections within each client, exacerbating the challenges in MFL.

The global model $F$ requires comprehensive processing of all modalities, precluding the discarding of any modality during local training.
This necessitates reconstruction of missing modalities $M'_i = M \setminus M_i$ through feature generation from available client data.
Existing reconstruction methods range from naive zero-filling \cite{Van_2012_Flexible} to advanced self-supervised synthesis \cite{Li_2017_Deep, Zong_2024_Self}. However, zero-filling frequently degrades model accuracy, while self-supervised methods that reduce label dependency fail to adequately address the complex challenge of cross-modal alignment \cite{Ma_2021_SMIL}.
Considering the limited storage and communication capacity of clients in MFL, it is crucial to adopt a lightweight and more efficient alignment method.

While some methods have focused on the alignment issues during the modality reconstruction process \cite{Xiong_2023_Client, Yu_2023_Multimodal, Wang_2024_FedMMR, Yan_2024_Cross}, they inadequately preserve semantic consistency across samples sharing identical labels. As demonstrated in Figure~\ref{fig:problems}, reconstructed modalities show inconsistent feature representations despite matching labels. This arises because existing methods align modalities through label correspondence but ignore semantic nuances specific to individual samples.
Such limitations highlight that label agreement alone cannot guarantee multimodal coherence, as inherent variations within classes persist across clients\footnote{Related work is discussed in Appendix A.}.

In this paper, we present \textbf{FedRecon}, a novel framework designed to address missing modality reconstruction in heterogeneous distributed environments. By employing a lightweight Multimodal Variational Autoencoder (MVAE) \cite{Kingma_2014_VAE, Wu_2018_MVAE, Shi_2019_MMVAE, Palumbo_2023_MMVAE+} as the core engine, FedRecon performs cross-modal alignment and modality imputation while overcoming client resource constraints in training and communication.
Compared to computationally intensive alternatives like GANs \cite{Goodfellow_2014_GAN} and Diffusion models \cite{Ho_2020_Diffusion}, our architecture substantially reduces training time and communication cost through its streamlined encoder-decoder design, with complete efficiency investigation provided in \textbf{Appendix B}.
Critically, we introduce a latent vector mapping mechanism that uniquely establishes sample-level alignment through direct feature coupling across modalities, ensuring semantic consistency during reconstruction.
Furthermore, we maintain a frozen auxiliary model initialized from the global MVAE to stabilize the generation process in Non-IID environments, collaborating with local models to generate missing modalities.

The contributions of this paper are as follows:
\begin{itemize}
    \item We are the first to apply MVAEs for realistic partial modality reconstruction in MFL under the setting where modality absence is randomly simulated based on missing rates, reflecting real-world scenarios.
    \item We propose the first latent variable mapping mechanism for MVAE-based reconstruction, achieving sample-level alignment without dependency on joint sampling. Additionally, we are also the first to introduce a global generator freezing strategy in MFL, effectively mitigating catastrophic forgetting in Non-IID environments.
    \item Experimental results validate state-of-the-art performance, surpassing baseline methods by 3.06\% and 5.28\% absolute gains on MELD and CrisisMMD benchmarks.
    Quality evaluations on CUB and PolyMNIST further demonstrate the effectiveness of FedRecon.
\end{itemize}

\section{Problem Formulation}

We consider a standard MFL framework involving \(K\) clients, where the complete modality set $M$ is heterogeneously distributed across clients. Each client $i$ possesses a private dataset $D_i = \{(X_i^j, y_i^j)\}_{j=1}^{n_i}$, where \(X_i^j = \{ X_{i,m}^j \in \mathbb{R}^{d_{x_m}} \ | \ m \in M_i \}\) represents the multimodal inputs and \(y_i^j \in Y_i\) denotes the corresponding label. Here, \(M_i \subseteq M\) indicates the subset of modalities available to client \(i\), \(d_{x_m}\) specifies the feature dimension of modality \(m\), and \(n_i = |D_i|\) is the number of data points available at client \(i\). The label space \(Y_i\) follows a client-specific distribution that determines both the label categories and their cardinality within the local dataset. The set of missing modalities for client \(i\) is denoted by \(M'_i = M \setminus M_i\).

Our goal is to train a global model \(F\) that includes modality-specific feature extractors \(\{ E_i : \mathbb{R}^{d_{x_m}} \to \mathbb{R}^{d_{f_m}} \}_{i \in M}\), where \(d_{f_m}\) is the feature space for modality \(m\), along with a shared classifier \(H\).
The global model undergoes initialization and dissemination to all clients at the beginning of each communication round \(t \in \{1, \ldots, T\}\). During round \(t\), every client \(i \in \{1, \ldots, K\}\) executes local model training utilizing their respective available modalities \(M_i\).
For the missing modalities \(M'_i\), reconstructing the missing data is essential to maintaining consistency in local training across clients, and is achieved using a generator \(G\).

Thus, the local loss function for client \(i\) on the sample \((X_i, y_i)\) can be expressed as:
\begin{equation}
\mathcal{L}_i(X_i, y_i; \phi_i) = \mathcal{L}_{\text{task}}\left(H\left( \oplus_{m \in M_i} \{ f_{i,m} \} \right), y_i \right),
\end{equation}
where \(f_{i,m} = E_{i,m}(X_{i,m})\) is the feature extracted by encoder \(E_{i,m}\) for modality \(m\), and \(\oplus\) denotes the fusion of available modalities.  
Following multi-task MFL frameworks \cite{Wang_2024_FedMMR, Feng_2023_Fedmultimodal}, the global objective is:
\begin{equation}
\min_{\phi} \mathcal{L}(F(\phi)) = \frac{1}{K} \sum_{i=1}^{K} \mathcal{L}_i(F_i(X_i, y_i; \phi_i)),
\end{equation}
where \(\phi_i\) denotes the set of parameters for client \(i\) and \(F_i\) represents the client-specific model trained on \(M_i\).

\section{Proposed Method}

\begin{figure*}[ht]
    \centering
    \includegraphics[width=\linewidth]{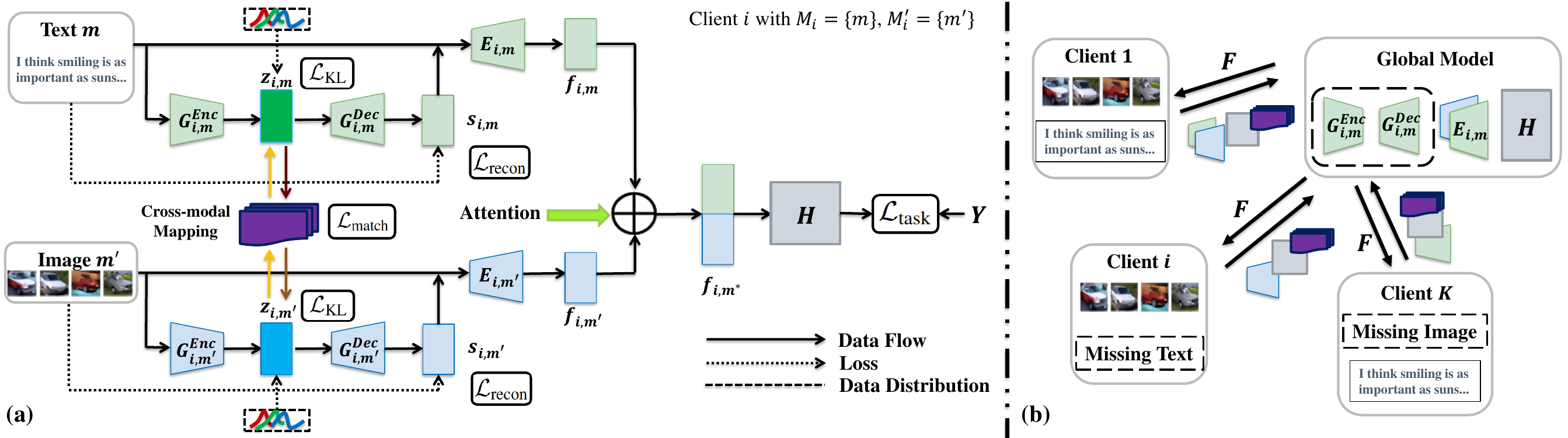}
    \caption{(a) The workflow of FedRecon, illustrating the process of reconstructing missing modality in distributed
    heterogeneous environments. (b) The communication mechanism of FedRecon, where each client downloads and submits model according to its modalities, reducing overhead during distributed learning.}
    \label{fig:workflow}
\end{figure*}

A promising approach to mitigate the missing modality issue is to enforce consistency across clients \cite{Wang_2024_FedMMR}. Our method builds on this foundation with two key components: (1) a lightweight MVAE that reconstructs missing modalities via an innovative distribution mapping mechanism to preserve cross-modal consistency, and (2) a global generator freezing strategy to stabilize Non-IID learning dynamics. The complete workflow is shown in Figure~\ref{fig:workflow}.

\subsection{\textbf{Missing Modality Reconstruction}}

Considering the challenges of training cost, storage capacity, and communication overhead in MFL scenarios, we conducted a comprehensive survey of various generative architectures. Among them, VAE emerges as particularly advantageous due to its relatively compact parameter structure. This characteristic translates to diminished computational requirements and storage consumption on client devices, while simultaneously minimizing data transmission burdens. Such attributes render VAE exceptionally suitable for MFL where efficient resource utilization is critical.

Each modality \( m \in M \) has its own VAE-based generator \( \text{G}_{i,m} \) for data reconstruction, which includes two main components: the encoder \( G_{i,m}^{\text{enc}} \) and the decoder \( G_{i,m}^{\text{dec}} \). The encoder maps the input \( X_{i,m} \) to a distribution over the latent space, while the decoder generates data \( X_{i,m} \) from the latent variable \( z \). These generators and local models are uploaded to the server, from where they are shared with other clients in the next MFL communication round. 

The goal of VAE is to maximize the log-likelihood of the data, but due to computational complexity, it is often optimized via a lower bound \cite{Kingma_2014_VAE}. The optimization objective of VAE is:
\begin{equation}
\label{original_formula}
\begin{aligned}
&\mathcal{L}_{\text{G}_{i,m}}(X_{i,m}) = -\mathbb{E}_{q_{\psi}(z|X_{i,m})} \left[ \log \frac{p_{\theta}(X_{i,m}, z)}{q_{\psi}(z|X_{i,m})} \right] \\
&= - \mathbb{E}_{q_{\psi}(z|X_{i,m})} \left[ \log p_{\theta}(X_{i,m} | z) \right] + D_{\text{KL}} \left( q_{\psi}(z|X_{i,m}) \| p(z) \right),
\end{aligned}
\end{equation}
where the input data \( X_{i,m} \) corresponds to modality \( m \), with \( z \) denoting the latent variables. The approximate posterior \( q_{\psi}(z|X_{i,m}) \) is parameterized by \( G_{i,m}^{\text{enc}} \) with learnable parameters \( \psi \), and \( p_{\theta}(X_{i,m}, z) = p(z)p_{\theta}(X_{i,m}|z) \) is the joint distribution generated by \( G_{i,m}^{\text{dec}} \) with parameters \( \theta \).

As a result, the loss can be divided into two components. The reconstruction loss \( \mathcal{L}_{\text{recon}} \) quantifies the ability of the decoder to accurately reconstruct the input data \( X_{i,m} \) based on the latent variable \( z \), whereas the KL divergence term \( \mathcal{L}_{\text{KL}} \) regularizes the posterior distribution \( q_{\psi}(z|X_{i,m}) \), encouraging it to approximate the prior distribution \( p(z) \). Notably, the VAE-based generator we adopt is label-agnostic. By leveraging the label space \( Y_i \), the generator models a distribution that aligns with the client’s attributes, serving as a high-level abstraction of the raw data and mitigating privacy risks without exposing sensitive information.

\subsection{\textbf{Cross-modal Alignment and Modality Awareness Fusion}}
\label{sec:Cross-modal Alignment and Modality Awareness Fusion}

Merely performing self-reconstruction using the data of the missing modality is insufficient. Leveraging cross-modal alignment to guide the generator is indispensable. Considering the issue of cross-modal interactions, the reconstructed modality and the other available modalities should exhibit similarity in the feature space. This fundamental insight drives our expansion of generator capabilities from self-reconstruction to cross-modal synthesis.

While advanced MVAE architectures leverage shared latent variables \cite{Shi_2019_MMVAE, Palumbo_2023_MMVAE+} to enable cross-modal interaction, they often suffer from limited diversity in reconstructed outputs and fail to achieve precise sample-level alignment across modalities. Our framework introduces a distribution re-encoding mechanism through mapping models \( T_{n,m} \), which adapt the latent variable \( z_m \) to better suit each target modality \( n \).
This design provides modality-specific latent inputs for each decoder, allowing for both decoupled distribution modeling and more accurate sample-level alignment across modalities.
The optimization objective in the multimodal setting becomes:
\begin{equation}
\label{final_formula}
\begin{aligned}
&\mathcal{L}_{\text{G}_{i,m}}(X_{i}) = - \frac{1}{M} \sum_{m=1}^{M} \mathbb{E}_{q_{\psi_m}(z_m|X_{i,m})} \left[ \log \frac{p_\Theta(X_{i}, z_m)}{q_{\psi_m}(z_m|X_{i,m})} \right], \\
&p_\Theta(X_{i}, z_m) = p(z_m) \, p_{\theta_m}(X_{i,m} | z_m) \prod_{\substack{n=1 \\ n \neq m}}^{M} p_{\theta_n}(X_{i,n} | T_{n,m}(z_m)),
\end{aligned}
\end{equation}
where \( T_{n,m} \) is a mapping model that adjusts the distribution of \( z_m \) to approximate \( z_n \), effectively bridging the representational gap between heterogeneous modalities.

Compared to optimizing feature alignment across different modalities using similarity at the feature level \cite{Wang_2024_FedMMR, Liu_2023_LLaVa}, directly aligning a compressed distribution proves to be more efficient. The structure of the model \( T_{n,m} \) is based on a simple MLP architecture, which projects the latent distribution from modality \( m \) onto that of modality \( n \) in the latent space.
However, we found that directly training the model using the objective in Equation~\ref{final_formula} leads to instability and poor convergence. To address this, we decompose the training process into two stages.
In the first stage, we jointly train the encoder and decoder components using the objective in Equation~\ref{original_formula} to learn modality-specific representations and enable accurate reconstruction.
In the second stage, we freeze the encoder and decoder parameters and exclusively optimize \( T_{n,m} \) so that the latent variable encoded from modality \( m \) can be transformed to faithfully reconstruct the data from modality \( n \).

To further stabilize this training process, we optionally apply KL divergence as a regularization term to promote convergence when \( T_{n,m} \) is hard to optimize. This encourages the mapped distribution to align better with the target latent space. The final training objective for \( T_{n,m} \) is defined as:
\begin{equation}
\mathcal{L}_{\text{match}} = \gamma \cdot D_{\text{KL}}\left( T_{n,m}(z_m) \,\|\, z_n \right) + \mathcal{L}_{\text{recon}},
\end{equation}
where \( \gamma \in \{0, 1\} \) is a binary coefficient controlling the use of the KL regularization term.

To enhance semantic consistency between the reconstructed and available modalities, we adopt an attention-based feature fusion mechanism inspired by \cite{Feng_2023_Fedmultimodal}. Given the extracted features \( f_{i,m} \) and its reconstructed counterpart \( f_{i,m}' \), we concatenate them into a joint representation \( h = [f_{i,m}; f_{i,m}'] \), and compute attention weights as follows:
\begin{equation}
\begin{aligned}
u = \tanh(W h + &b), \quad a = \text{softmax}(u^T c), \\
{\phantom{\sum_i}} f_{i,m}^* &= \sum_i a_i h_i,
\end{aligned}
\end{equation}
where \( W \), \( b \), and \( c \) are learnable parameters, and the final fused feature \( f_{i,m}^* \) is obtained through a weighted sum over the concatenated input. This attention mechanism can be extended to a multi-head setting by introducing multiple context vectors, enabling finer-grained fusion across modalities.

\subsection{\textbf{Global Generator Freezing Strategy}}

In Section~\ref{sec:Cross-modal Alignment and Modality Awareness Fusion}, we train the generator using complete modality data \( X^{j}_{i} \), and employ it during inference to predict the missing modality \( X^{k}_{i,m'} \) from the available modality \( X^{k}_{i,m} \).
However, in Non-IID settings, the missing modality \( X^{k}_{i,m'} \) may exhibit a noticeable distribution shift with respect to the local data of client \( i \), making it underrepresented during local training and thus prone to forgetting. In such cases, the missing data \( X^{k}_{i,m'} \) may be better aligned with the distribution of another client \( i' \), where similar samples \( X^{j}_{i'} \) are more common. Consequently, relying solely on local generators can lead to suboptimal reconstruction performance.

\begin{figure}[t]
    \centering
    \fbox{\includegraphics[width=0.45\textwidth]{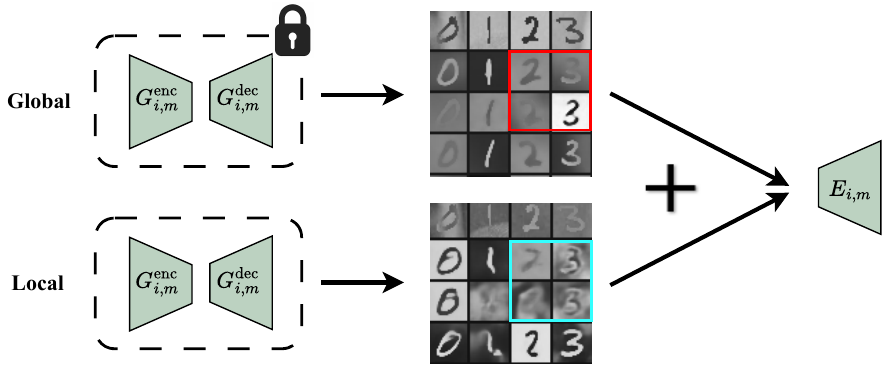}}
    \caption{An illustrative example of the Global Generator Freezing Strategy on the PolyMNIST dataset.}
    \label{fig:communication_mechanism}
\end{figure}

To mitigate this, we propose a \textit{global generator freezing strategy}, where each client maintains two generators: one global generator initialized at deployment and kept frozen, and another updated locally. This allows clients to fall back on the global generator when their own reconstruction suffers from data shift or forgetting.
As illustrated in Figure~\ref{fig:communication_mechanism}, we replace the original synthetic data with a concatenation of outputs from both global and local generators, and reduce the weight of synthetic features during the attention fusion process to achieve a balancing effect.
Despite the additional requirement of storing both generators and mapping models on each client, our approach remains significantly more lightweight than GAN- or Diffusion-based alternatives. The detailed algorithmic workflow is outlined in \textbf{Appendix C}.

\section{\textbf{Experiments}}

\subsection{\textbf{Experimental Setups}}

\begin{figure*}[t]
    \centering
    \begin{subfigure}{0.24\textwidth}
        \centering
        \includegraphics[width=\textwidth]{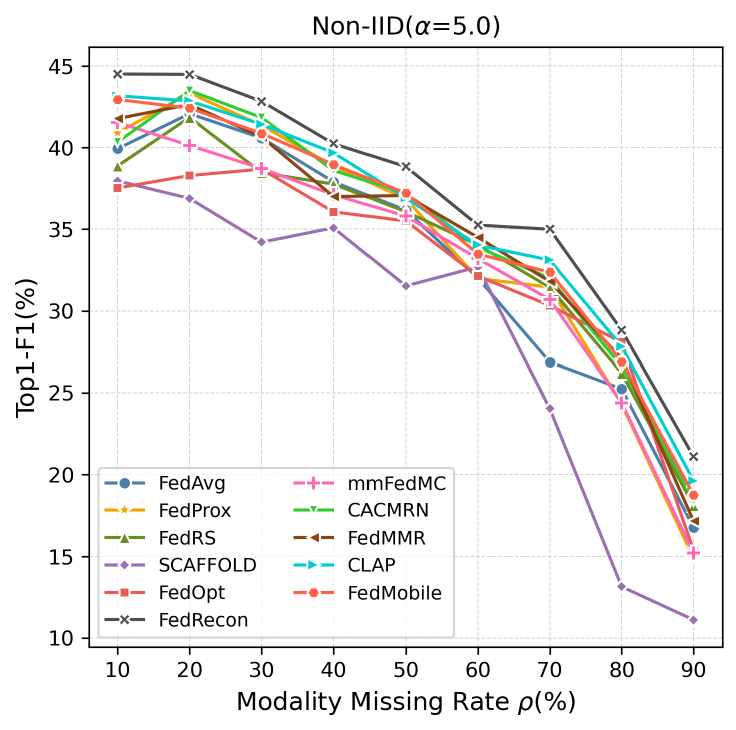}
        \caption{Top1-F1, $\alpha=5.0$}
    \end{subfigure}
    \begin{subfigure}{0.24\textwidth}
        \centering
        \includegraphics[width=\textwidth]{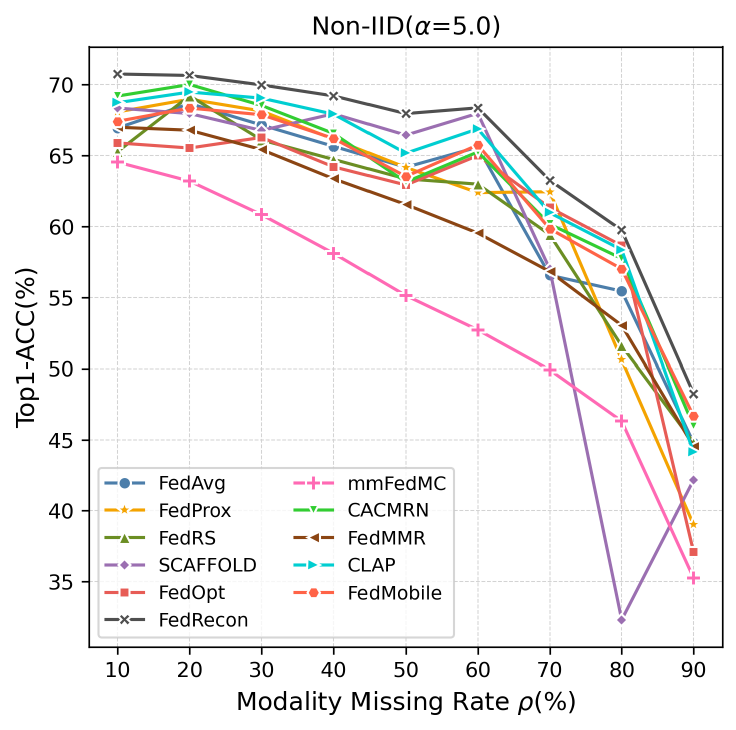}
        \caption{Top1-accuracy, $\alpha=5.0$}
    \end{subfigure}
    \begin{subfigure}{0.24\textwidth}
        \centering
        \includegraphics[width=\textwidth]{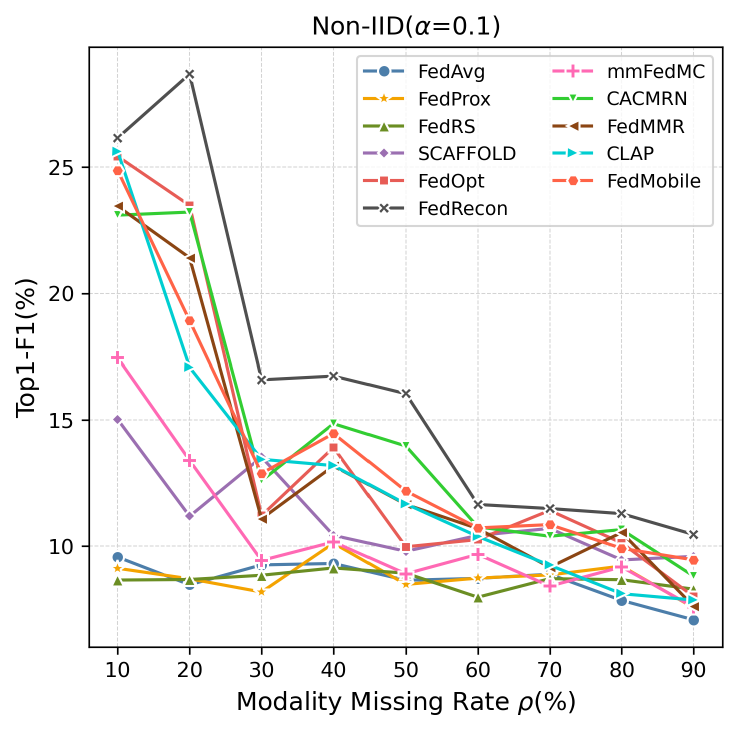}
        \caption{Top1-F1, $\alpha=0.1$}
    \end{subfigure}
    \begin{subfigure}{0.24\textwidth}
        \centering
        \includegraphics[width=\textwidth]{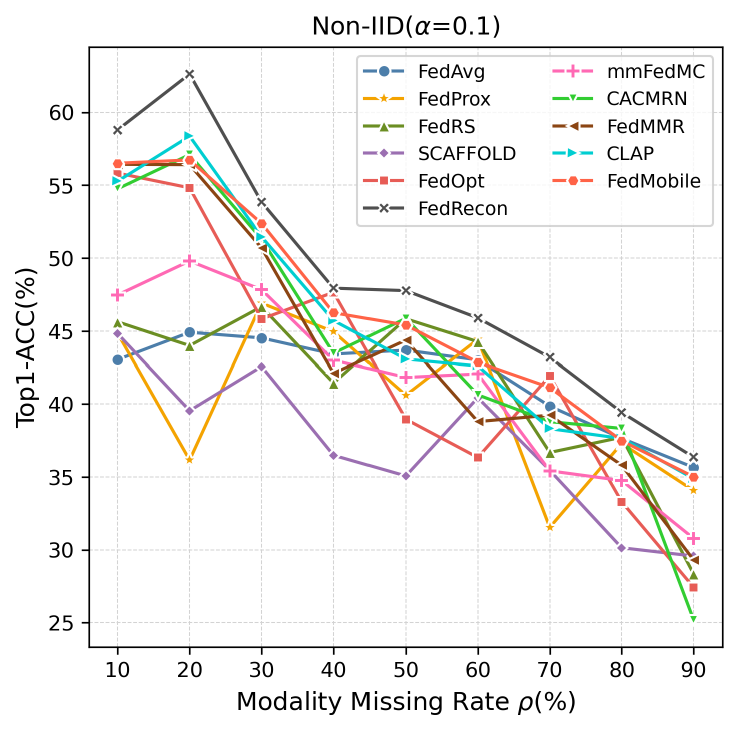}
        \caption{Top1-accuracy, $\alpha=0.1$}
    \end{subfigure}
    \caption{Top1-F1 and Top1-accuracy on CrisisMMD under Dedicated FL scenarios, comparing the effects of low ($\alpha=0.1$) and high ($\alpha=5.0$) label imbalance.}
    \label{fig:result}
\end{figure*}

\begin{table}[t]
\centering
\renewcommand{\arraystretch}{1.1}
\setlength{\tabcolsep}{1.8pt}
{\small
\resizebox{1.05\linewidth}{!}{
\begin{tabular}{l|cccccccccccc}
\hline
& \multicolumn{9}{c}{\textbf{MELD, Top1-Unweighted Average Recall (UAR) $\uparrow$}} \\ \hline
\textbf{Algorithm} & \textbf{10\%} & \textbf{20\%} & \textbf{30\%} & \textbf{40\%} & \textbf{50\%} & \textbf{60\%} & \textbf{70\%} & \textbf{80\%} & \textbf{90\%} \\ \hline
FedAvg & 55.03 & 53.62 & 54.04 & 53.11 & 52.73 & 51.53 & 50.60 & 48.31 & 46.65\\
FedProx & 54.78 & 53.49 & 53.04 & 53.09 & 53.08 & 51.65 & 50.47 & 49.16 & 48.45\\
FedRS & 54.24 & 53.19 & 53.81 & 52.80 & 53.26 & 51.39 & 51.79 & 49.52 & 46.53\\
SCAFF. & 50.87 & 49.82 & 48.49 & 48.14 & 47.16 & 46.34 & 42.87 & 40.43 & 29.44\\
FedOpt & 54.67 & 52.38 & 51.88 & 51.69 & 52.81 & 49.93 & 48.54 & 49.78 & 47.94\\
\hline
CACMRN & 54.62 & 53.44 & 54.43 & 54.03 & 53.58 & 52.31 & 50.56 & 50.23 & 47.57\\
mmFedMC & 55.31 & 53.30 & 53.85 & 50.95 & 52.65 & 50.13 & 50.03 & 48.98 & 46.10\\
FedMMR & 54.47 & 53.49 & 52.83 & 52.78 & 54.06 & 51.52 & 50.69 & 49.26 & 47.79\\
CLAP & 55.24 & 55.09 & \underline{55.34} & \underline{54.81} & 53.90 & 51.84 & 50.12 & 48.18 & 45.18\\
FedMobile & \underline{55.86} & \underline{55.42} & 54.83 & 54.49 & \underline{54.52} & \underline{54.26} & \textbf{53.94} & \underline{52.91} & \underline{52.08}\\
FedRecon & \textbf{56.53} & \textbf{56.18} & \textbf{55.95} & \textbf{55.78} & \textbf{55.31} & \textbf{54.90} & \underline{53.85} & \textbf{53.53} & \textbf{52.84}\\ 
\hline
\end{tabular}
}
}
\caption{Comparison of UAR classification results for Dedicated FL on MELD, with $\rho$ varying from 10\% to 90\%.}
\label{tab:MELD_UAR}
\end{table}

\begin{table}[t]
\centering
\renewcommand{\arraystretch}{1.1}
\setlength{\tabcolsep}{1.8pt}
{\small
\resizebox{1.05\linewidth}{!}{
\begin{tabular}{l|ccccccccc}
\hline
& \multicolumn{8}{c}{\textbf{MELD, Top1-accuracy $\uparrow$}} \\ \hline
\textbf{Algorithm} & \textbf{10\%} & \textbf{20\%} & \textbf{30\%} & \textbf{40\%} & \textbf{50\%} & \textbf{60\%} & \textbf{70\%} & \textbf{80\%} & \textbf{90\%} \\ \hline
FedAvg & 67.09 & 67.05 & 66.78 & 66.02 & 63.46 & 64.66 & 64.81 & 62.51 & 61.39\\
FedProx & 67.51 & 67.33 & 65.82 & 67.60 & 64.61 & 63.92 & 64.34 & 63.91 & 64.81\\
FedRS & 65.95 & 65.92 & 66.76 & 64.41 & 63.59 & 63.42 & 64.28 & 64.68 & 63.60\\
SCAFF. & 69.24 & \underline{68.79} & 68.64 & \underline{68.26} & 67.93 & 67.74 & 66.12 & 64.22 & 57.46\\
FedOpt & 66.70 & 66.42 & 64.89 & 65.73 & 65.69 & 60.20 & 61.03 & 62.98 & 61.01\\
\hline
CACMRN & 68.48 & 67.47 & 63.36 & 67.36 & 67.02 & \underline{67.97} & 65.18 & 64.76 & 63.64\\
mmFedMC & 65.90 & 66.40 & \underline{68.83} & 64.95 & 65.89 & 64.94 & 64.09 & 62.34 & 61.92\\
FedMMR & 68.26 & 67.13 & 68.09 & 65.89 & 67.38 & 63.47 & 66.17 & 65.09 & 64.71\\
CLAP & \underline{69.71} & 68.14 & 67.99 & 67.86 & 67.44 & 65.41 & 64.51 & 62.08 & 60.53\\
FedMobile & 69.55 & 68.18 & 67.91 & 67.45 & \underline{67.98} & 67.81 & \underline{67.10} & \underline{65.45} & \underline{64.94}\\
FedRecon & \textbf{70.83} & \textbf{69.51} & \textbf{69.67} & \textbf{69.16} & \textbf{68.86} & \textbf{68.76} & \textbf{68.13} & \textbf{67.58} & \textbf{66.41}\\ 
\hline
\end{tabular}
}
}
\caption{Comparison of Top1-acc classification results for Dedicated FL on MELD, with $\rho$ varying from 10\% to 90\%.}
\label{tab:MELD_ACC}
\end{table}

\textbf{Datasets.} To evaluate our MFL framework, we utilize two real-world multimodal datasets: \textbf{(1) MELD} \cite{Poria_2018_Meld}, a multiparty conversational dataset comprising over 9k utterances with both audio and textual transcripts from the TV series Friends, and \textbf{(2) CrisisMMD} \cite{Alam_2018_CrisisMMD}, which contains 18.1k tweets accompanied by both visual and textual information, collected during various crisis events.
To further assess the generative quality of our MVAE model, we additionally adopt two benchmark multimodal datasets: \textbf{(3) PolyMNIST} \cite{Sutter_2021_PolyMNIST}, a synthetic dataset featuring five modalities, where each consists of MNIST digits composited over diverse background images; and \textbf{(4) CUB} \cite{Wah_2011_CUB}, the Caltech-UCSD Birds dataset, which pairs bird images with fine-grained textual descriptions, posing a challenging setting due to the modality-specific nuances in visual and textual representations.

\textbf{Baselines.} We compare FedRecon against a range of both unimodal and multimodal FL methods. Unimodal baselines include FedAvg~\cite{McMahan_2017_FedAvg}, FedProx~\cite{Li_2020_FedProx}, FedRS~\cite{Li_2021_FedRS}, Scaffold~\cite{Karimireddy_2020_Scaffold}, and FedOpt~\cite{Reddi_2020_FedOpt}, all implemented with zero-filling for missing modalities following the FedMultimodal~\cite{Feng_2023_Fedmultimodal} benchmark. Multimodal baselines include CACMRN \cite{Xiong_2023_Client}, FedMMR~\cite{Wang_2024_FedMMR}, mmFedMC~\cite{Yuan_2024_Communication}, CLAP~\cite{Cui_2024_CLAP} and FedMobile~\cite{Liu_2025_FedMobile} where zero-filling is likewise adopted in the absence of modality-specific generative modules. To assess the generative performance of our MVAE model, we compare against several representative multimodal generative approaches, including MVAE~\cite{Wu_2018_MVAE}, MVTCAE~\cite{Hwang_2021_MVTCAE}, mmJSD~\cite{Sutter_2020_mmJSD}, MMVAE~\cite{Shi_2019_MMVAE}, MoPoE-VAE~\cite{Sutter_2021_MoPoE-VAE}, MMVAE+~\cite{Palumbo_2023_MMVAE+}, and MMVM~\cite{Sutter_2024_MMVM}.

\begin{figure*}[t]
    \centering
    \begin{subfigure}{0.19\textwidth}
        \centering
        \includegraphics[width=\textwidth]{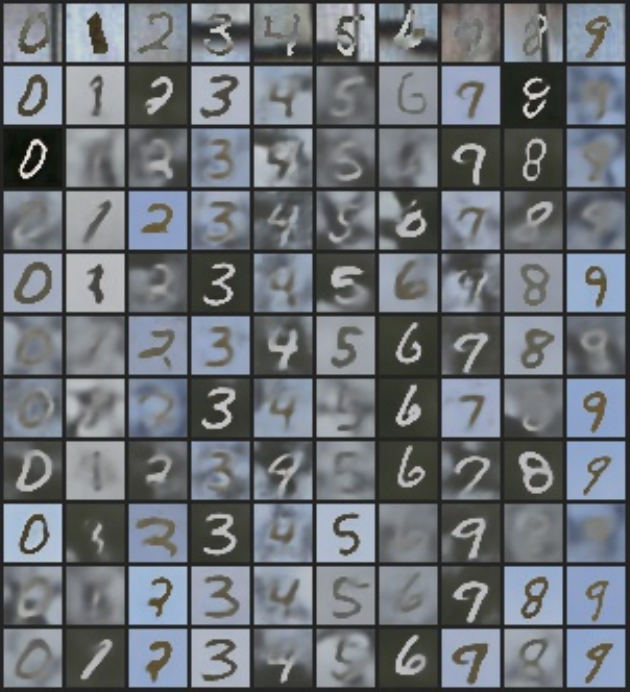}
        \caption{Generated m0 from m4}
    \end{subfigure}
    \begin{subfigure}{0.19\textwidth}
        \centering
        \includegraphics[width=\textwidth]{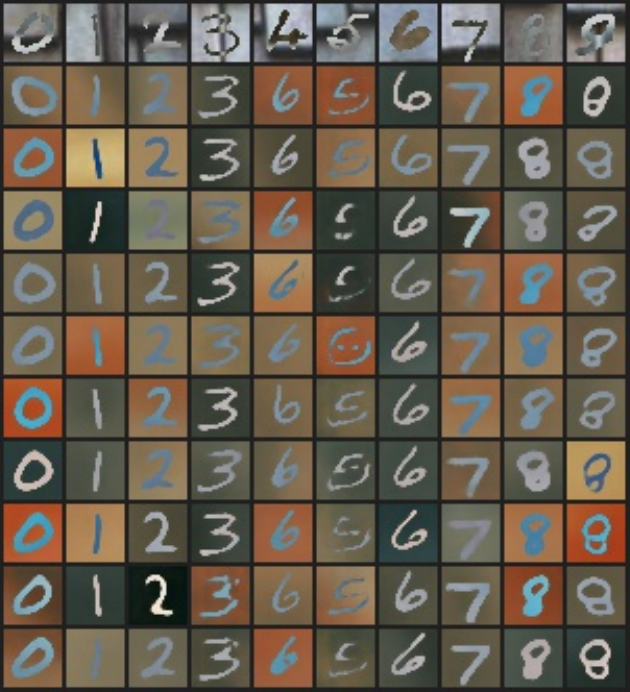}
        \caption{Generated m1 from m4}
    \end{subfigure}
    \begin{subfigure}{0.19\textwidth}
        \centering
        \includegraphics[width=\textwidth]{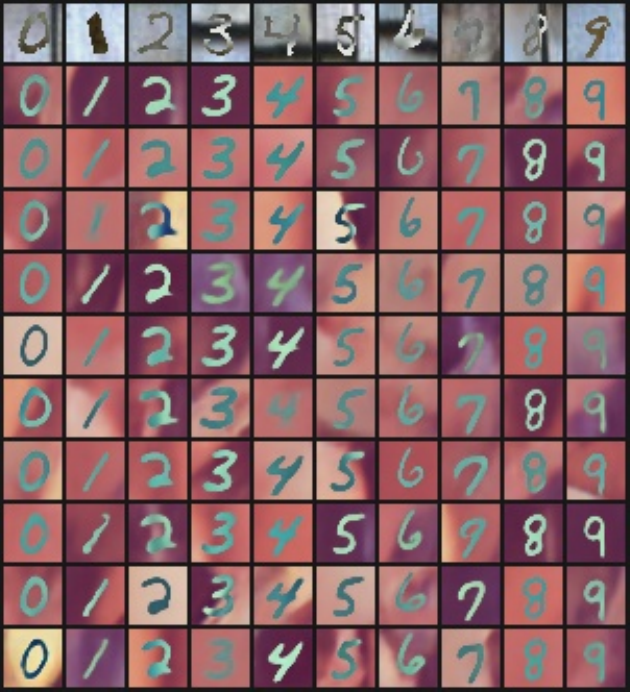}
        \caption{Generated m2 from m4}
    \end{subfigure}
    \begin{subfigure}{0.19\textwidth}
        \centering
        \includegraphics[width=\textwidth]{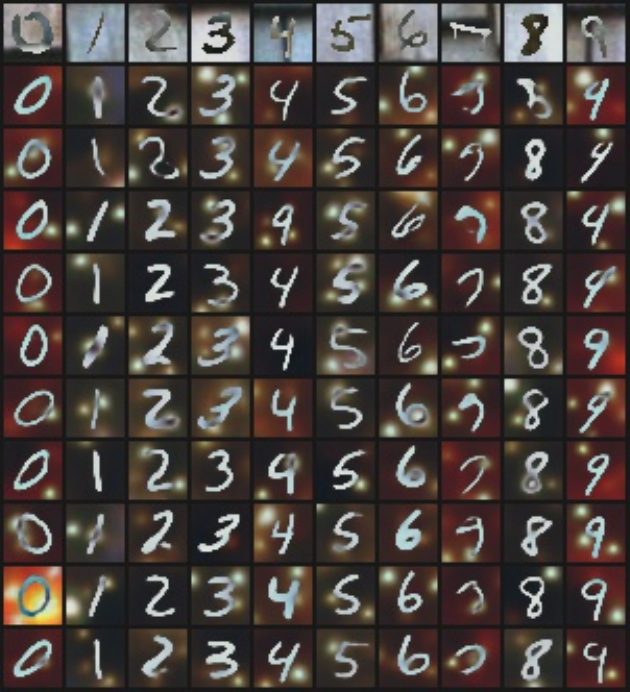}
        \caption{Generated m3 from m4}
    \end{subfigure}
    \begin{subfigure}{0.19\textwidth}
        \centering
        \includegraphics[width=\textwidth]{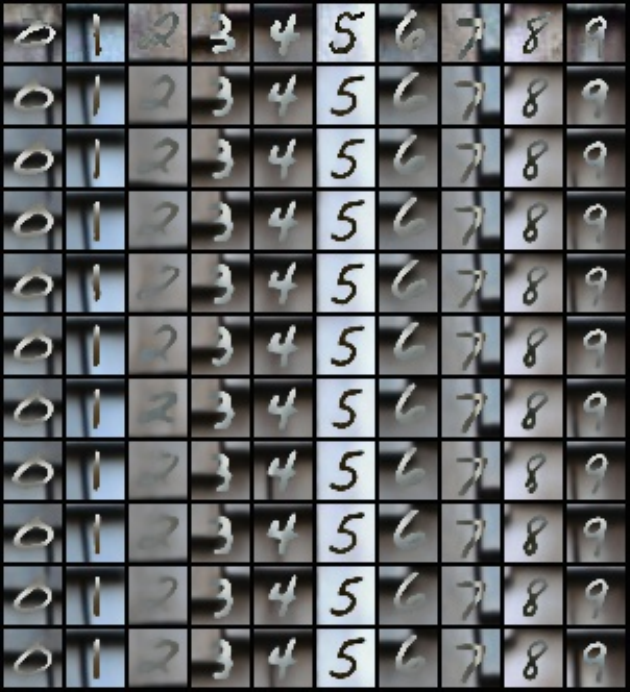}
        \caption{Generated m4 from m4}
    \end{subfigure}
    \caption{Cross-modal generation results on PolyMNIST. Each image is generated from modality m4, showcasing reconstructions and translations into modalities m0–m4.}
    \label{fig:cross_modal_generation}
\end{figure*}

\begin{figure*}[t]
    \centering
    \includegraphics[width=\textwidth]{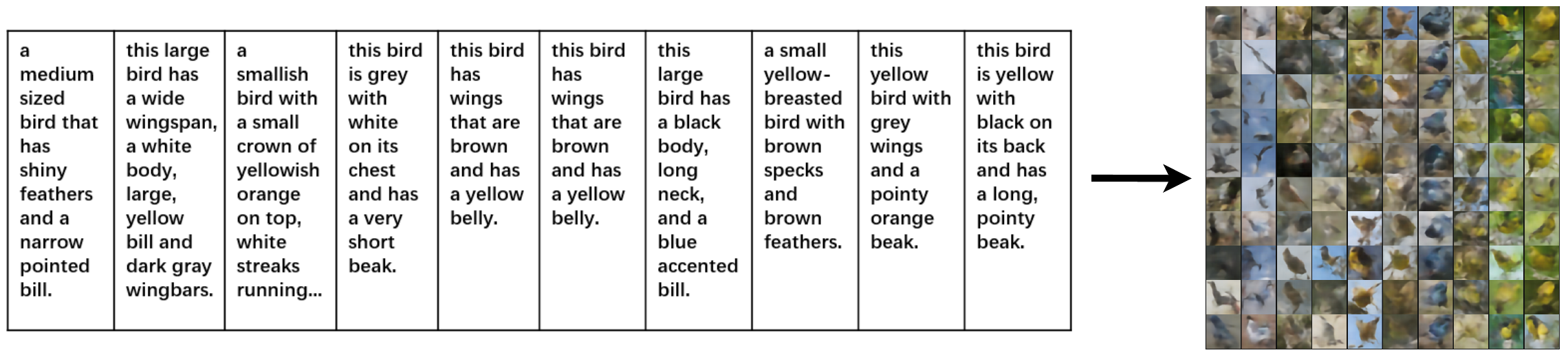}
    \caption{Cross-modal generation on CUB: images generated from the text modality (m1) to the image modality (m0). Each column corresponds to a set of images generated from the same text input.}
    \label{fig:cubTI}
\end{figure*}

\textbf{Implementations.} For our MFL framework, we implement FedRecon atop the FedMultimodal benchmark~\cite{Feng_2023_Fedmultimodal}. We use FedAvg as the baseline method for MELD, while for CrisisMMD we adopt FedOpt.
We report Unweighted Average Recall (UAR) on MELD and F1-score on CrisisMMD, along with Top-1 accuracy on both datasets for a more comprehensive evaluation.
To evaluate generative baselines, we build upon the official MMVAE+ implementation\footnote{https://github.com/epalu/mmvaeplus} and refer to their reported metrics for consistency. Generative quality is assessed using Fréchet Inception Distance (FID) \cite{Heusel_2017_FID} and generative coherence \cite{Shi_2019_MMVAE} metrics.
All experiments are repeated with three random seeds to ensure robustness and are conducted on a single NVIDIA A100 GPU with 80GB memory.
Further implementation details, including those for reproducing closed-source baselines, are provided in \textbf{Appendix E}.

\textbf{Modality Heterogeneity.} We simulate missing modalities by modeling their availability using a Bernoulli distribution, where each modality has an equal probability of being absent. In this work, we adopt the Dedicated FL paradigm \cite{Wang_2024_FedMMR}, where both the training and testing sets experience the same missing rate. The data missing rate is varied from 0.1 to 0.9.

\textbf{Statistical Heterogeneity.} For the MELD dataset, we partition the data based on client IDs, as the data is naturally organized by speaker identifiers. In contrast, the CrisisMMD dataset lacks such realistic client partitions, so we use the Dirichlet distribution to partition the data with \( \alpha \in \{0.1, 5.0\} \), where \( \alpha = 0.1 \) and \( \alpha = 5.0 \) correspond to high and low data heterogeneity, respectively.

\subsection{Main Results}

We evaluate FedRecon in two parts. The first part focuses on classification performance within the MFL setting, while the second part assesses generative quality based on the evaluation protocol from MMVAE+. To highlight performance differences, we use \textbf{boldface} to indicate the best result and \underline{underline} for the second-best.

\textbf{MELD and CrisisMMD Results.}
The MELD dataset consists of audio and text modalities, which convey highly consistent information. As a result, the impact of missing modalities on performance is relatively mild, and reconstruction brings limited gains.
As shown in Table~\ref{tab:MELD_UAR} and Table~\ref{tab:MELD_ACC}, FedRecon achieves strong performance, e.g., 55.31\% UAR at 50\% missing modalities, outperforming FedAvg by 2.58\%.
Notably, even under the extreme condition where 90\% of modalities are missing, it achieves a Top1-UAR of 52.84\% and a Top1-accuracy of 66.41\%.
Similarly, due to sample distribution issues, we evaluate CrisisMMD using both F1-score and Top1-accuracy. Performance declines as the missing rate increases across all methods. Despite this inherent fluctuations, FedRecon consistently outperforms its counterparts, achieving an average F1-score improvement of 2.81\% over FedOpt at high imbalance ($\alpha=0.1$) and 4.34\% under low imbalance ($\alpha=5.0$).

\begin{table}[t]
\centering
\renewcommand{\arraystretch}{1.1}
\setlength{\tabcolsep}{2.0pt}
{\small
\begin{tabular}{l|cc|cc}
\hline
\multirow{2}{*}{\textbf{Method}} & \multicolumn{2}{c|}{\textbf{PolyMNIST}} & \multicolumn{2}{c}{\textbf{CUB}} \\
& \textbf{Coherence} & \textbf{FID} & \textbf{Coherence} & \textbf{FID} \\
\hline
MVAE & 0.093 \tiny({$\pm$0.009}) & 82.59 \tiny({$\pm$6.22}) & 0.271 \tiny({$\pm$0.007}) & 172.21 \tiny({$\pm$39.61}) \\
MVTCAE & 0.509 \tiny({$\pm$0.006}) & \textbf{58.98 \tiny({$\pm$0.62})} & 0.221 \tiny({$\pm$0.007}) & 208.43 \tiny({$\pm$1.10}) \\
mmJSD & 0.785 \tiny({$\pm$0.023}) & 209.98 \tiny({$\pm$1.26}) & 0.556 \tiny({$\pm$0.158}) & 262.80 \tiny({$\pm$6.93}) \\
MMVAE & 0.837 \tiny({$\pm$0.004}) & 152.11 \tiny({$\pm$4.11}) & 0.713 \tiny({$\pm$0.057}) & 232.20 \tiny({$\pm$2.14}) \\
MoPoE. & 0.723 \tiny({$\pm$0.006}) & 160.29 \tiny({$\pm$4.12}) & 0.579 \tiny({$\pm$0.158}) & 265.55 \tiny({$\pm$4.01}) \\
MMVAE+ & 0.796 \tiny({$\pm$0.010}) & 80.75 \tiny({$\pm$0.18}) & 0.721 \tiny({$\pm$0.090}) & 164.94 \tiny({$\pm$1.50}) \\
\hline
MMVM & 0.773 \tiny({$\pm$0.041}) & 98.34 \tiny({$\pm$7.11}) & 0.714 \tiny({$\pm$0.079}) & \textbf{160.44 \tiny({$\pm$3.20})} \\
FedRecon & \textbf{0.845 \tiny({$\pm$1.199})} & 78.915 \tiny({$\pm$8.33}) & \textbf{0.729 \tiny({$\pm$0.062})} & 171.35 \tiny({$\pm$10.45}) \\
\hline
\end{tabular}
}
\caption{Conditional generation performance on PolyMNIST and CUB. Higher coherence and lower FID indicate better performance. We adopt the default hyperparameters used in MMVAE+ for fair comparison.}
\label{MVAE_Result}
\end{table}

\textbf{PolyMNIST and CUB Results.}
As shown in Table~\ref{MVAE_Result}, we directly adopt the reported results of MMVAE+ as a reference benchmark, and further compare FedRecon against MMVM for a comprehensive evaluation.
In terms of FID, FedRecon performs relatively poorly on the CUB dataset but yields satisfactory results on PolyMNIST. We speculate that this discrepancy stems from the difference in latent distribution characteristics: the Laplace distribution used for CUB is more challenging to map effectively, whereas the Gaussian distribution adopted for PolyMNIST aligns more naturally with our mapping mechanism.
While the reconstruction quality is constrained by this setup, our method achieves sample-level alignment across modalities, which helps improve the overall data distribution and leads to a slight FID improvement over the original MVAE.
Regarding generative coherence, since we achieve sample-level alignment, the accuracy of the classifier improves significantly, yielding superior results. FedRecon outperforms both MMVAE+ and MMVM, reaching state-of-the-art performance in this domain.
Figure~\ref{fig:cross_modal_generation} and Figure~\ref{fig:cubTI} showcase several randomly selected cross-modal generation results on PolyMNIST and CUB test sets.
More results can be found in \textbf{Appendix F}.

\begin{figure}[t]
    \centering
    \includegraphics[width=0.45\textwidth]{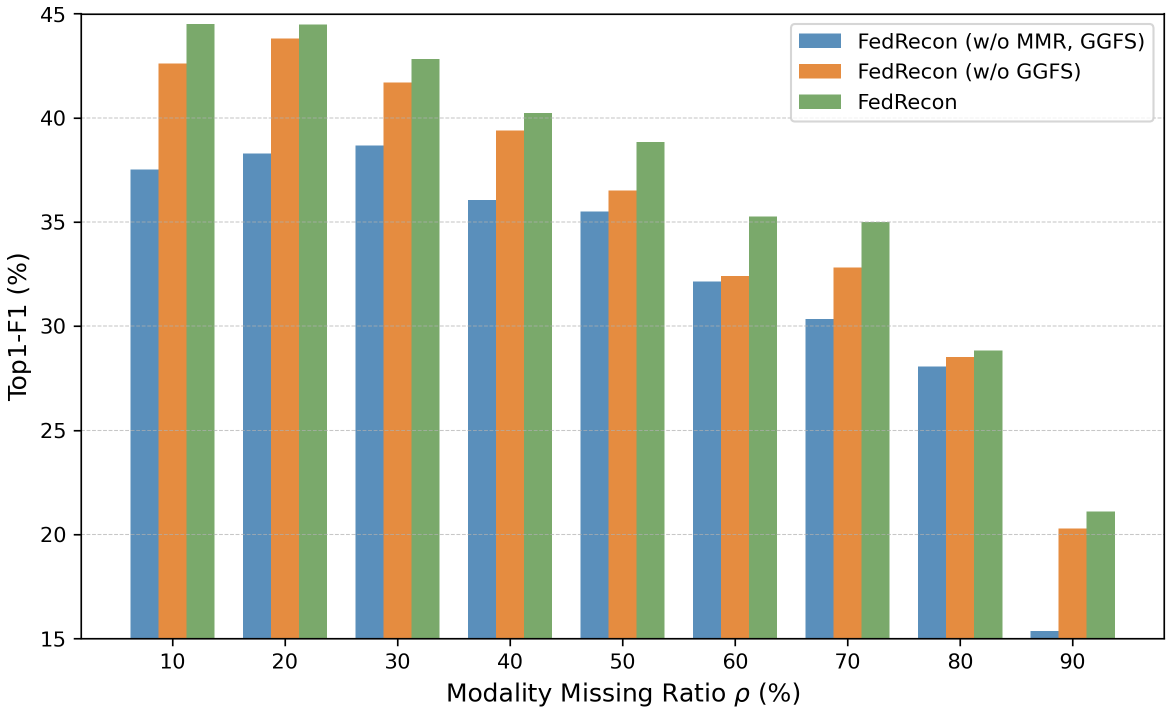}
    \caption{Ablation results on CrisisMMD under $\alpha=5.0$.}
    \label{fig:ablation}
\end{figure}

\subsection{\textbf{Ablation Studies}}

\indent

\textbf{Contribution of Each Module.} We evaluate the impact of two key modules: the cross-modal alignment reconstruction generator (MMR) and the Global Generator Freezing Strategy (GGFS). To assess their contributions to model performance, we conduct experiments where these modules are excluded from FedRecon. The results underscore the critical role of these components in enhancing overall effectiveness. As shown in Figure~\ref{fig:ablation}, FedRecon demonstrates a marked improvement over the baseline, highlighting the necessity of both MMR and GGFS for superior performance.
Owing to space limitations, detailed results on CrisisMMD with $\alpha=0.1$ and MELD are provided in \textbf{Appendix G}. Nonetheless, we briefly summarize the findings here.
On CrisisMMD ($\alpha=0.1$), MMR and GGFS achieve average F1 gains of 2.79\% and 1.55\% respectively, across varying missing rates. For MELD, GGFS shows limited impact, with average gains around 0.49\%, while the MMR module consistently improves UAR by 2.03\% on average.

\textbf{Impact of Modality-Aware Fusion.}
This component can be universally applied to all comparison methods by adaptively weighting the modalities before classification. To ensure a fair comparison in reconstructing missing modalities, we incorporate this mechanism across all baselines. Ablating this module yields divergent effects across datasets: on MELD, the UAR score increases by 5\%–6\%, whereas on CrisisMMD, the F1-score decreases by 3\%–4\%. These outcomes align with the observations reported in FedMultimodal, reflecting the nuanced influence of modality-aware fusion depending on dataset characteristics.

\textbf{Rationale Behind GGFS.} We hypothesize that certain clients hold samples inherently difficult to reconstruct due to limited local data coverage, reflecting scenarios where local distributions fail to capture global complexity. Supporting evidence for the hypothesis, in the form of label distribution visualizations, along with additional analysis of the GGFS module and qualitative comparisons of generated samples with and without GGFS, are provided in \textbf{Appendix H}.

\subsection{Further Analysis}

\indent

\textbf{Training Efficiency.} We provide a detailed training cost analysis in \textbf{Appendix I}, showing that FedRecon remains efficient even with a large number of modalities.

\textbf{Mapping Complexity Reduction.} We can reduce complexity from \(O(M^2)\) to \(O(M)\) by assuming each function \(T_{n,m}\) is not injective and maps any modality \( m \)’s latent vectors to \( n \)’s space. Thus, only one per \(T_{n}\) modality is needed.
Nevertheless, even with the original \(T_{n,m}\), FedRecon trains 80\% faster than the full MVAE on PolyMNIST when \(|M|=5\). The evaluation of the simplified design, which shows only a very slight performance decrease, is given in \textbf{Appendix J}.

\textbf{Extensibility to Arbitrary VAE Frameworks}. Our mapping mechanism is extensible to any pre-trained VAE by freezing its encoder and decoder and learning to align latent distributions across modalities. We apply it to MMVAE+, with corresponding generation results on CUB presented in \textbf{Appendix K}. Additionally, high-resolution images (e.g., CrisisMMD) pose challenges for VAE-based generation. This can be eased by scaling up the VAE or using diffusion models when resources permit.

\textbf{Comparison with CLAP}. CLAP is the method most similar to ours, but differences in its original setup lead to suboptimal results. Details are discussed in \textbf{Appendix L}.

\section{\textbf{Conclusion}}

In this paper, we tackle the challenge of missing modalities in the MFL paradigm, which often leads to feature distribution gaps and disrupted inter-modal correlations. We propose FedRecon, a novel framework that incorporates two key components: (1) a cross-modal reconstruction strategy that enables both label-level and sample-level imputation, and (2) a global generator freezing strategy designed to address the issue of unreconstructable samples in Non-IID settings with limited data.
Extensive experiments across diverse scenarios demonstrate the effectiveness of FedRecon in enhancing both modality reconstruction and overall performance under Non-IID conditions, paving the way for more robust and practical MFL systems.

{\small
\bibliography{aaai2026}
}

\newpage
\clearpage

\appendix

\section{Related Work}
\subsection{Multimodal Variational Autoencoder}

The Multimodal Variational Autoencoder (MVAE) extends the framework of Variational Autoencoders (VAE) \cite{Kingma_2014_VAE} to enable joint learning and generation across heterogeneous data modalities. Wu and Goodman \cite{Wu_2018_MVAE} pioneered this direction by introducing a scalable architecture capable of learning shared representations in weakly supervised settings, where only a subset of data samples contain all modalities. Subsequent work further advanced the generative coherence and fidelity of MVAEs. Shi \textit{et al.} \cite{Shi_2019_MMVAE} proposed a mixture-of-experts formulation (MMVAE) that decomposes latent spaces into shared and modality-specific subspaces while ensuring coherent joint generation through multimodal integration. Palumbo \textit{et al.} \cite{Palumbo_2023_MMVAE+} enhanced this framework by explicitly isolating distributions of shared and private latent factors, thereby improving robustness to variations in latent space design without compromising semantic consistency. Further innovations integrated Diffusion models into the MVAE architecture \cite{Palumbo_2024_Deep}, demonstrating enhanced unconditional generation performance for complex real-world data modalities.
Despite their effectiveness in aligning modalities through joint latent space learning, current methods exhibit limitations in maintaining consistent cross-modal alignment across diverse samples sharing the same label, especially when used on highly variable datasets or data with real-world noise.
Notably, FedRecon mitigates these deficiencies through a simple mapping mechanism, demonstrating practical efficacy in cross-modal alignment.

\subsection{Multimodal Federated Learning}

Multimodal Federated Learning (MFL) addresses privacy-preserving collaboration across clients with heterogeneous modality types in real-world applications. Ouyang \textit{et al.} propose Harmony \cite{Ouyang_2023_Harmony}, which disentangles training into modality-wise federated learning and fusion learning, leveraging balance-aware resource allocation to improve accuracy and speed convergence in Non-IID scenarios. Gao \textit{et al.} develop FedMVD \cite{Gao_2025_Fusion}, introducing global logit alignment and local angular margin mechanisms to mitigate domain shifts and category imbalance across multimodal nodes. Chen and Zhang present FedMBridge \cite{Chen_2024_FedMBridge}, using a topology-aware hypernetwork to bridge statistical and architectural heterogeneity for personalized multimodal architectures efficiently.
While these approaches advance MFL under modality heterogeneity, they generally assume the availability of all modalities on clients. Our method, FedRecon, focuses on missing modality reconstruction in heterogeneous distributed environments, stabilizing training via a frozen global generator replica to reduce distributional shifts caused by client-specific data heterogeneity, thus enhancing usability and robustness in practical scenarios.

\subsection{Missing Modality Reconstruction in MFL}

Reconstructing missing modalities in MFL ensures robust model performance under heterogeneous data distributions. Xiong \textit{et al.} propose CACMRN \cite{Xiong_2023_Client}, a client-adaptive transformer leveraging instance relationships and federated optimization to prevent local overfitting during cross-modal reconstruction. Wang \textit{et al.} develop FedMMR \cite{Wang_2024_FedMMR}, aligning feature spaces via modality synthesis while integrating reconstructed and existing modalities as complementary inputs. Liu \textit{et al.} design FedMobile \cite{Liu_2025_FedMobile}, prioritizing cross-node latent feature sharing to address extreme modality incompleteness. Saha \textit{et al.} investigate modality incongruity issues in healthcare FL and explore modality imputation and regularization techniques to reduce missing modality effects \cite{Saha_2025_Incongruent}. Yu \textit{et al.} introduce FedInMM \cite{Yu_2024_Robust}, utilizing LSTM-based dynamic weighting and temporal fusion for irregular multimodal data.
While effective in reconstructing modalities under limited samples in MFL, these methods overlook sample-level alignment across clients with identical labels.
Additionally, they inadequately address instability caused by Non-IID client data during cross-modal generation.
FedRecon stabilizes training by freezing a global generator replica to reduce distributional shifts, thereby mitigating fluctuations induced by client-specific data divergence.

\section{A Survey on Generative Models}

To assess the parameter sizes of various models, we conducted an extensive survey on platforms, including \textbf{GitHub}, \textbf{Hugging Face}, \textbf{Papers with Code}, focusing on representative architectures from three major types of generative models: GANs, VAEs, and Diffusion models. Specifically, we considered ACGAN \cite{Odena_2017_ACGAN}, DeGAN \cite{Addepalli_2020_DeGAN}, Seg\_DeGAN \cite{bhogale2020data}, MADGAN \cite{Ghosh_2018_CVPR}, MAD-GAN \cite{li2019mad}, BEGAN \cite{berthelot2017began}, Boundary-Seeking GAN \cite{hjelm2017boundary}, StyleGAN2 \cite{Karras_2020_CVPR} and DragGAN \cite{10.1145/3588432.3591500} from the GAN class, VAE \cite{Kingma_2014_VAE}, VQ-VAE \cite{Oord_2017_Neural}, VAR \cite{Tian_2024}, MVAE \cite{Wu_2018_MVAE} and MMVAE+ \cite{Palumbo_2023_MMVAE+} from the VAE class, as well as DDPM \cite{Ho_2020_Diffusion}, DDPM-Seg \cite{baranchuk2021label}, Tab-DDPM \cite{kotelnikov2023tabddpm}, Stable Diffusion \cite{rombach2022high} and Mini-EDM \cite{Karras_2022_Elucidating} from the Diffusion models. These models were selected for their prominence and diversity, and we sampled and computed their parameter sizes to facilitate a comparative analysis.

\begin{table}[htbp]
\centering
\scalebox{0.9}{
\begin{threeparttable}
\caption{Comparison of Model Parameters and File Sizes}
\label{tab:model_comparison}
\begin{tabular}{llcc}
\toprule
\textbf{Category} & \textbf{Model} & \textbf{Parameters (M)} & \textbf{Size (MB)} \\
\midrule
\multirow{9}{*}{GAN} 
    & ACGAN\tnote{1} & 6.64 & 24.53 \\
    & DeGAN\tnote{2} & 7.15 & 24.17 \\
    & Seg-DeGAN\tnote{3} & 7.15 & 24.19 \\
    & MADGAN\tnote{4} & 0.34 & 1.36 \\
    & MAD-GAN\tnote{5} & 1.05 & 3.96 \\
    & BEGAN\tnote{6} & 1.79 & 7.15 \\
    & B-S GAN\tnote{7} & 1.96 & 7.82 \\
    & DragGAN\tnote{8} & 59.26 & 347.17 \\
    & StyleGAN2\tnote{9} & 59.26 & 347.10 \\
\midrule
\multirow{6}{*}{VAE} 
    & Simple VAE\tnote{10} & 0.21 & 0.80 \\
    & VQ-VAE\tnote{11} & 0.26 & 1.00 \\
    & VAR\tnote{12} & 310.28 & 1185.55 \\
    & VAE in MVAE & 4.43 & 17.72 \\
    & MVAE & 22.16 & 84.63 \\
    & MMVAE+\tnote{13} & 22.16 & 84.63 \\
    
\midrule
\multirow{5}{*}{Diffusion} 
    & DDPM\tnote{14} & 35.87 & 136.85 \\
    & DDPM-Seg\tnote{15} & 2.20 & 8.41 \\
    & Tab-DDPM\tnote{16} & 1.44 & 5.77 \\
    & Stable Diffusion\tnote{17} &  1068.40 & 4372.48 \\
    & Mini-EDM\tnote{18} & 15.71 & 59.92 \\
\bottomrule
\end{tabular}

\begin{tablenotes}
\footnotesize
\item[1] \url{https://github.com/lukedeo/keras-acgan}
\item[2] \url{https://github.com/vcl-iisc/DeGAN}
\item[3] \url{https://github.com/kaushal-py/seg-degan}
\item[4] \url{https://github.com/vinx-2105/MADGAN-DEGAN}
\item[5] \url{https://github.com/Guillem96/madgan-pytorch}
\item[6] \url{https://github.com/eriklindernoren/PyTorch-GAN}
\item[7] \url{https://github.com/eriklindernoren/PyTorch-GAN}
\item[8] \url{https://github.com/XingangPan/DragGAN}
\item[9] \url{https://github.com/NVlabs/stylegan2}
\item[10] \url{https://github.com/bojone/vae}
\item[11] \url{https://github.com/MishaLaskin/vqvae}
\item[12] \url{https://github.com/FoundationVision/VAR}
\item[13] \url{https://github.com/epalu/mmvaeplus}
\item[14] \url{https://github.com/zoubohao/DenoisingDiffusionProbabilityModel-ddpm-}
\item[15] \url{https://github.com/yandex-research/ddpm-segmentation}
\item[16] \url{https://github.com/yandex-research/tab-ddpm}
\item[17] \url{https://github.com/CompVis/stable-diffusion}
\item[18] \url{https://github.com/yuanzhi-zhu/mini_edm}

\end{tablenotes}
\end{threeparttable}
}
\end{table}

As illustrated in Table~\ref{tab:model_comparison}, for repositories containing multiple model variants, we selected the smallest model for comparison to ensure fairness in parameter and size analysis. VAE exhibits a lightweight advantage over the other two model families, delivering competitive performance while maintaining significantly lower resource consumption. However, VAE also has its limitations. When adapted to multimodal tasks, the quality of the synthesized data often falls short compared to other generative models, exhibiting noticeable artifacts such as blurry edges in reconstructions~\cite{Vivekananthan_2024_Comparative, Kazerouni_2022_Diffusion}.
Regarding training time, since each model is designed for different tasks and configurations, a unified evaluation is challenging. Nevertheless, we conducted a comparative test using the most basic implementations of VAE, GAN, and diffusion models on the CIFAR-10 \cite{krizhevsky2009learning} dataset. Among them, VAE demonstrated the fastest training speed.

\section{Pseudocode}

The pseudocode of our method is given in Algorithm~\ref{algorithm}.

\begin{algorithm}[t]
\caption{Two-Stage Federated Training for FedRecon}
\label{algorithm}
\textbf{Notations:} 
$\phi^{(t)}$: global model parameters at round $t$; 
$G_{k,m}^{(t)}$: MVAE parameters (encoder and decoder) for client $k$ and modality $m$; 
$\widetilde{G}_{k,m}^{(t)}$: frozen global MVAE parameters;
$T_{n, m}^{(t)}$: mapping model from modality $m$ to $n$, abbreviated as $T^{(t)}$;
$\widetilde{T}_{n,m}^{(t)}$: frozen mapping model

\begin{algorithmic}[1]
\FOR{round $t = 1$ to $T$}
    \FOR{client $k = 1$ to $K$ \textbf{in parallel}}
        \STATE \textbf{Stage 1: Local training of MVAE}
        \STATE $\{\widetilde{G}_{k,m}^{(t)}\}_{m=1}^M, \{G_{k,m}^{(t)}\}_{m=1}^M \gets \{G_{m}^{(t-1)}\}_{m=1}^M$
        \STATE $\widetilde{T}_k^{(t)}, T_k^{(t)} \gets T^{(t-1)}$
        \STATE Update $\{G_{k,m}^{(t)}\}_{m=1}^M$ using $\mathcal{L}_{\text{G}_{k,m}}$ for $L_1$ epochs

        \STATE \textbf{Stage 2: Freeze MVAE and train mapping modules}
        \STATE $\{G_{k,m}^{(t)}\}_{m=1}^M \gets \text{Freeze}(\{G_{k,m}^{(t)}\}_{m=1}^M)$
        \STATE Update $T_k^{(t)} \gets \text{SGD}(T_k^{(t)})$ for $L_2$ epochs

        \STATE \textbf{Stage 3: Update task model}
        \STATE Reconstruct using both global $\widetilde{G}_{k,n}^{(t)}$ and local $G_{k,n}^{(t)}$
        \STATE Update $\phi_k^{(t)} \gets \text{SGD}(\phi^{(t-1)})$ for $L_3$ epochs
    \ENDFOR

    \STATE Aggregate:
    \STATE $\{G_{m}^{(t)}\}_{m=1}^M \gets \frac{1}{K}\sum_{k=1}^K G_{k,m}^{(t)}$
    \STATE $T^{(t)} \gets \frac{1}{K}\sum_{k=1}^K T_k^{(t)}$
    \STATE $\phi^{(t)} \gets \frac{1}{K}\sum_{k=1}^K \phi_k^{(t)}$
\ENDFOR
\end{algorithmic}
\end{algorithm}

\section{Datasets}

\textbf{Datasets.} We employ four multimodal datasets in our experiments, serving two distinct purposes. To evaluate the effectiveness of our multimodal federated learning framework, we use two real-world datasets: \textbf{(1) MELD} \cite{Poria_2018_Meld}, a multiparty conversational corpus comprising over 9k utterances with both audio and textual transcripts from the TV series Friends, and \textbf{(2) CrisisMMD} \cite{Alam_2018_CrisisMMD}, which contains 18.1k tweets with paired visual and textual information collected during various real-world crisis events.
To further assess the generative quality of our MVAE model, we additionally adopt two benchmark multimodal datasets: \textbf{(3) PolyMNIST} \cite{Sutter_2021_PolyMNIST}, a synthetic dataset featuring five modalities, each constructed by compositing MNIST digits over diverse background images; and \textbf{(4) CUB} \cite{Wah_2011_CUB}, the Caltech-UCSD Birds dataset, which pairs bird images with fine-grained textual descriptions, posing a challenging setting due to the nuanced modality-specific alignments.

Next, we provide additional details on how the four datasets were utilized in our experiments. Our experimental setup is primarily based on the federated configuration from \cite{Feng_2023_Fedmultimodal}. For the MELD dataset, we applied a simplification by retaining only the four most frequent emotion categories: neutral, happy, sad, and angry, in order to reduce class imbalance and improve model convergence. For CrisisMMD, we adopted the original configuration and partitioned the data following the event-based distribution protocol described in prior work. Detailed statistics of these two datasets are provided in Tables~\ref{tab:MELD_statistics} and~\ref{tab:CrisisMMD_statistics}.

As for the remaining two datasets, PolyMNIST and CUB, we followed the standard processing protocols established in previous work \cite{Sutter_2021_PolyMNIST, Shi_2019_MMVAE, Wah_2011_CUB}.
PolyMNIST is a synthetic dataset constructed from five image modalities. Each sample comprises five distinct images, each representing an MNIST digit placed over a randomly cropped background from five unique texture images—one per modality. All five images share the same digit label, which constitutes the shared information across modalities, while the style of handwriting and the background image vary across modalities, capturing modality-specific noise. The digit labels serve as the class labels in our experiments.
CUB (Caltech-UCSD Birds) is a fine-grained dataset that pairs bird images with natural language descriptions. Each datapoint includes an image of a bird and one or more corresponding captions. This dataset poses a more realistic and challenging multimodal learning scenario due to the substantial amount of modality-specific information: the image conveys visual appearance details (e.g., plumage patterns), while the captions vary in focus and granularity, leading to varying degrees of semantic overlap between modalities.

\begin{figure*}[t]
    \centering
    \fbox{\includegraphics[width=0.95\textwidth,height=0.28\textwidth]{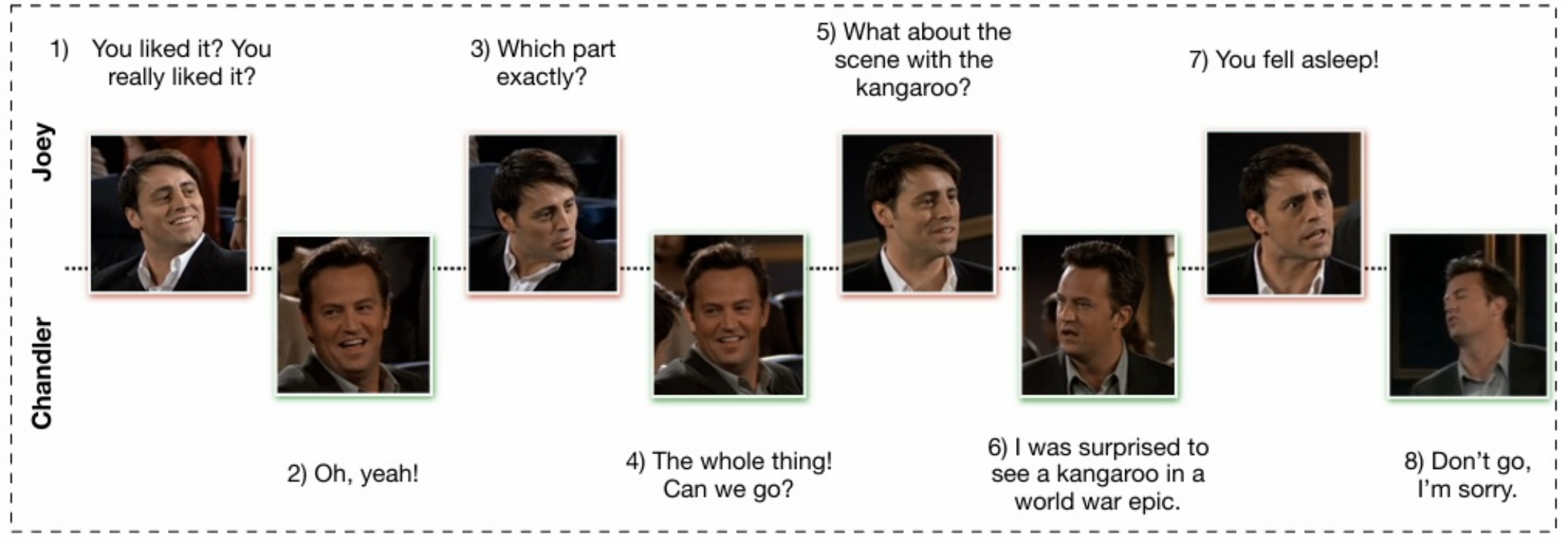}}
    \caption{An example from the MELD dataset, illustrating multimodal data in emotional dialogue analysis.}
    \label{fig:MELD}
\end{figure*}

\begin{figure}[t]
    \centering
    \fbox{\includegraphics[width=0.45\textwidth,height=0.28\textwidth]{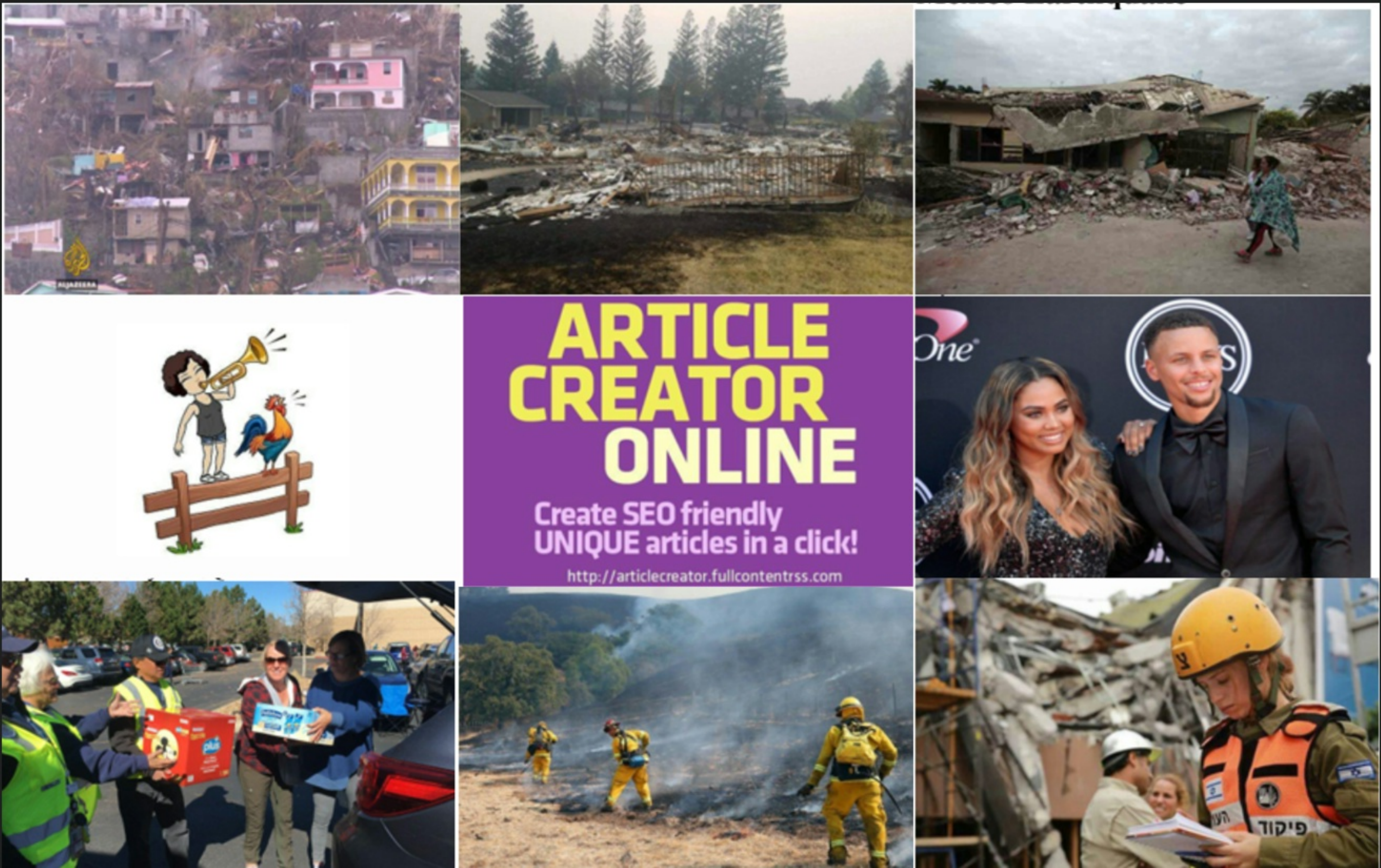}}
    \caption{An example from the CrisisMMD dataset, showing multimodal data used for crisis situation modeling.}
    \label{fig:CrisisMMD_A}
\end{figure}

\begin{figure}[t]
    \centering
    \fbox{\includegraphics[width=0.45\textwidth,height=0.28\textwidth]{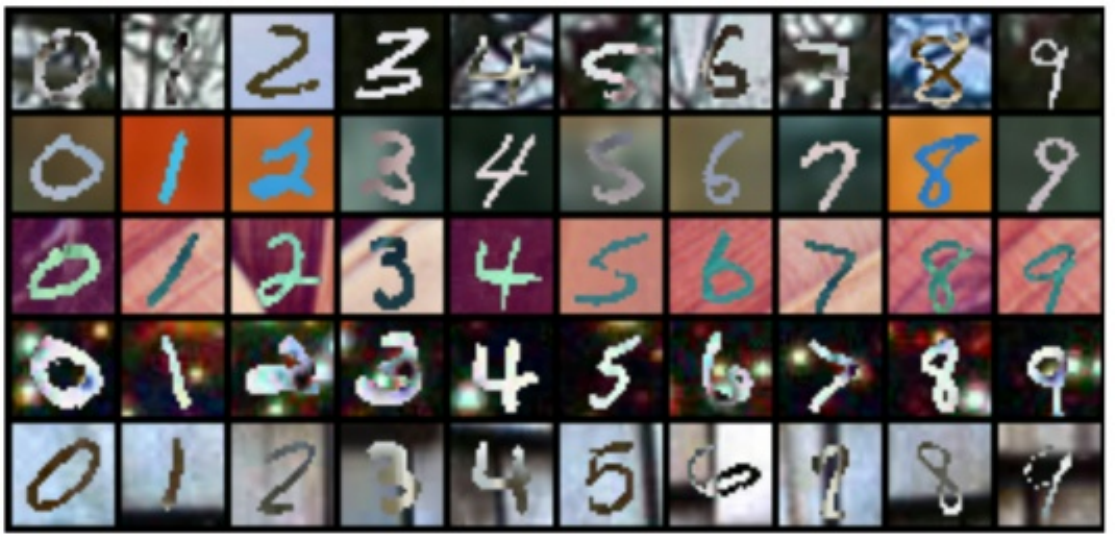}}
    \caption{An example from the PolyMNIST dataset, illustrating five synthetic modalities where each digit shares the same label but varies in background and handwriting style.}
    \label{fig:PolyMNIST}
\end{figure}

\begin{figure}[t]
    \centering
    \fbox{\includegraphics[width=0.45\textwidth,height=0.28\textwidth]{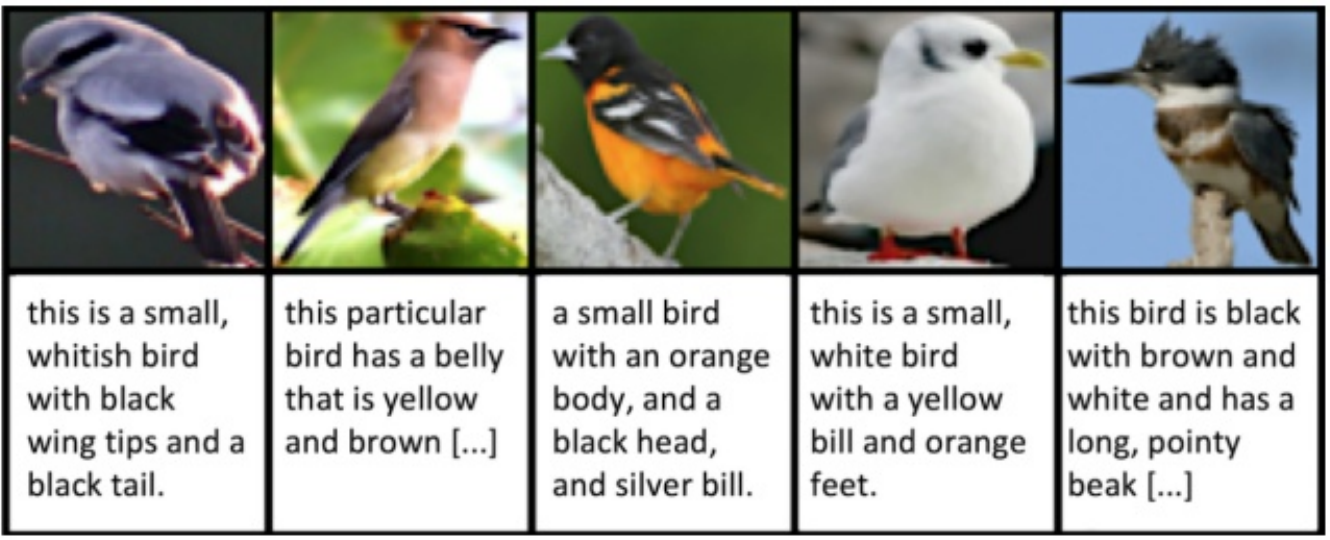}}
    \caption{An example from the CUB dataset, showing bird images paired with fine-grained textual descriptions, highlighting the modality-specific nuances in visual and linguistic features.}
    \label{fig:CUB}
\end{figure}

Together, these four datasets allow us to evaluate both the quality and robustness of generative models under varying conditions of modality alignment and shared content.
As illustrated in Figures~\ref{fig:MELD}, \ref{fig:CrisisMMD_A}, \ref{fig:PolyMNIST}, and~\ref{fig:CUB}, we present representative examples from the original datasets to better convey the nature and complexity of each modality. These examples help visualize the diversity of the multimodal inputs, as well as the varying levels of semantic overlap across modalities. The images are adapted from the MMVAE+ repository.

\begin{table}[t]
    \centering
    \caption{MELD Statistics. \{a, v, t\} = audio, visual, text}
    \label{tab:MELD_statistics}
    \scalebox{0.9}{
    \begin{tabular}{lccc}
        \toprule
        MELD Statistics & Train & Dev & Test \\
        \midrule
        \# of modalities & \{a,v,t\} & \{a,v,t\} & \{a,v,t\} \\
        \# of unique words & 10,643 & 2,384 & 4,361 \\
        Avg./Max utterance length & 8.0/69 & 7.9/37 & 8.2/45 \\
        \# of dialogues & 1039 & 114 & 280 \\
        \# of dialogues dyadic MELD & 2560 & 270 & 577 \\
        \# of utterances & 9989 & 1109 & 2610 \\
        \# of speakers & 260 & 47 & 100 \\
        Avg. \# of utterances per dialogue & 9.6 & 9.7 & 9.3 \\
        Avg. \# of emotions per dialogue & 3.3 & 3.3 & 3.2 \\
        Avg./Max \# of speakers per dialogue & 2.7/9 & 3.0/8 & 2.6/8 \\
        \# of emotion shift & 4003 & 427 & 1003 \\
        Avg. duration of an utterance & 3.59s & 3.59s & 3.58s \\
        \bottomrule
    \end{tabular}
    }
\end{table}

\begin{table}[t]
    \centering
    \caption{Crisis Dataset Statistics}
    \label{tab:CrisisMMD_statistics}
    \scalebox{0.70}{
    \begin{tabular}{lcccc}
        \toprule
        Crisis name & \# tweets & \# images & \# filtered tweets & \# sampled tweets \\
        \midrule
        Hurricane Irma & 3,517,280 & 176,972 & 5,739 & 4,041 (4,525) \\
        Hurricane Harvey & 6,664,349 & 321,435 & 19,967 & 4,000 (4,443) \\
        Hurricane Maria & 2,953,322 & 52,231 & 6,597 & 4,000 (4,562) \\
        California wildfires & 455,311 & 10,130 & 1,488 & 1,486 (1,589) \\
        Mexico earthquake & 383,341 & 7,111 & 1,241 & 1,239 (1,382) \\
        Iraq-Iran earthquake & 207,729 & 6,307 & 501 & 499 (600) \\
        Sri Lanka floods & 41,809 & 2,108 & 870 & 832 (1,025) \\
        \midrule
        Total & 14,223,141 & 576,294 & 36,403 & 16,097 (18,126) \\
        \bottomrule
    \end{tabular}}
\end{table}

\section{Implementation}

For our MFL framework, we implement FedRecon atop the FedMultimodal benchmark~\cite{Feng_2023_Fedmultimodal}. Following their protocol, we utilize MobileNetV2~\cite{Howard_2017_Mobilenets}, MobileBERT~\cite{Sun_2020_Mobilebert}, and MFCC (Mel-Frequency Cepstral Coefficients)~\cite{Davis_1980_MFCC} to extract features from visual, textual, and audio modalities respectively, enabling efficient and scalable federated training. Therefore, a simple Conv+RNN architecture is used to process visual and audio data, while text is handled by an RNN-only model.
For the MELD dataset, which contains only audio and text modalities, we adapt our model structure based on the VAE design provided in a prior implementation\footnote{\url{https://github.com/yjlolo/vae-audio}} to better handle the audio modality.
For CrisisMMD, we adopt the vision-language VAE architecture proposed in the original MVAE framework to accommodate the image-text modality pair.
We use FedAvg as the baseline method for MELD, while for CrisisMMD we adopt FedOpt.
We report Unweighted Average Recall (UAR) on MELD and F1-score on CrisisMMD, along with Top-1 accuracy on both datasets for a more comprehensive evaluation.
To evaluate generative baselines, we build upon the official MMVAE+ implementation\footnote{\url{https://github.com/epalu/mmvaeplus}} and refer to their reported metrics for consistency. We also reproduce MMVM\footnote{\url{https://github.com/thomassutter/mmvmvae}} results as an additional reference point. Generative quality is assessed using Fréchet Inception Distance (FID) \cite{Heusel_2017_FID} and generative coherence \cite{Shi_2019_MMVAE} metrics.
For the coefficient \(\gamma\) controlling the KL regularization term, we applied it in the experiments on the CIRSMMD and CUB datasets.
All experiments are repeated with three different random seeds to ensure robustness and are conducted on a single NVIDIA A100 GPU with 80GB memory.

For all task models, we adopt the training hyperparameter settings provided by the FedMultimodal benchmark to ensure fair comparison and reproducibility.  
We also detail the hyperparameters used for MVAE training, which include 20 local epochs per communication round, a learning rate of \(1 \times 10^{-3}\), and equal weighting coefficients of 1 for both the KL divergence loss (\(L_{\mathrm{KL}}\)) and the reconstruction loss (\(L_{\mathrm{recon}}\)). The mapping model is trained for 40 epochs per communication round with a learning rate of \(5 \times 10^{-3}\). Through empirical testing, we found these hyperparameter settings provide stable training performance and are well suited across all datasets used in our experiments.

Regarding the comparative experiments, all our unimodal baselines are based on implementations provided by FedMultimodal. For multimodal comparisons, we reused the first three methods—mmFedMC \cite{Yuan_2024_Communication}, CACMRN \cite{Xiong_2023_Client}, and FedMMR \cite{Wang_2024_FedMMR}—as none of them have publicly available code. For CLAP \cite{Cui_2024_CLAP} and FedMobile \cite{Liu_2025_FedMobile}, we based our implementation on their open-source repositories. In the following, we provide detailed explanations of how each of these baselines was implemented in our setup.

For \textbf{mmFedMC}, the original paper does not address the issue of modality completion, focusing instead on communication, aggregation, and client selection strategies. We followed the formulas and procedures described in the paper, preserving the ensemble model setup and uploading the task model for evaluation. Clients are ranked based on a combination of Shapley values and task loss, and the final model is aggregated accordingly.

For \textbf{CACMRN}, we adopted their Transformer-based reconstruction module. However, since FedMultimodal provides pre-extracted modality features, we directly applied the reconstruction mechanism to these features without incorporating an additional feature extractor.
Specifically, we implemented the \textit{Normalized Self-Attention (NSA)} mechanism as described in the original work. In NSA, both the input features and the projection matrices (\(W_Q, W_K, W_V\)) are L2-normalized, projecting them onto a hyperspherical space. This operation treats the parameters as cluster centers of the data distribution. Attention scores are computed between instances to adaptively update the projection matrices that are most relevant to the local data, thereby mitigating overfitting risks.
Clients with complete multimodal data train the reconstruction model, optimized using three loss functions: Cycle-Consistency Loss (\(L_{CM}\)) minimizes MSE between original and reconstructed features in both directions; Semantic Consistency Loss (\(L_{SCL}\)) aligns features by minimizing KL divergence in classification outputs; and Divergence Loss (\(L_{DI}\)) encourages diversity in projection matrices to enhance self-attention.
We followed the original experimental hyperparameters for all these components to ensure a faithful reproduction of CACMRN's reconstruction behavior within our federated setting.

As for \textbf{FedMMR}, the original paper did not specify the exact architecture used for its generator. Therefore, we implemented a generator based on standard GAN architectures and followed similar experimental settings. Since the original experiments were conducted on image and audio datasets and focused on reconstructing a single missing modality, we adopted the same setup in our reproduction.

For \textbf{CLAP} and \textbf{FedMobile}, we adjusted their generator architectures to better fit our data modalities and conducted distributed training under our experimental setup.
We categorized them into two types based on whether sequence data (e.g., text, audio) is present or not (e.g., images). For each category, we adopted the corresponding original architecture from their official implementations.
Ultimately, we evaluated the performance using our defined task model alongside the trained generators.

We have open-sourced the code for training the mapping models. You can adapt and use any pretrained MVAE model to work with our repository. Our MVAE training is based on MMVAE and MMVAE+, and you may refer to their open-source code for implementation details. Our federated learning testing is based on FedMultimodal, where you can download datasets and initialize models accordingly.

\section{More Rusults}

\begin{figure}[t]
    \centering
    \includegraphics[width=0.45\textwidth]{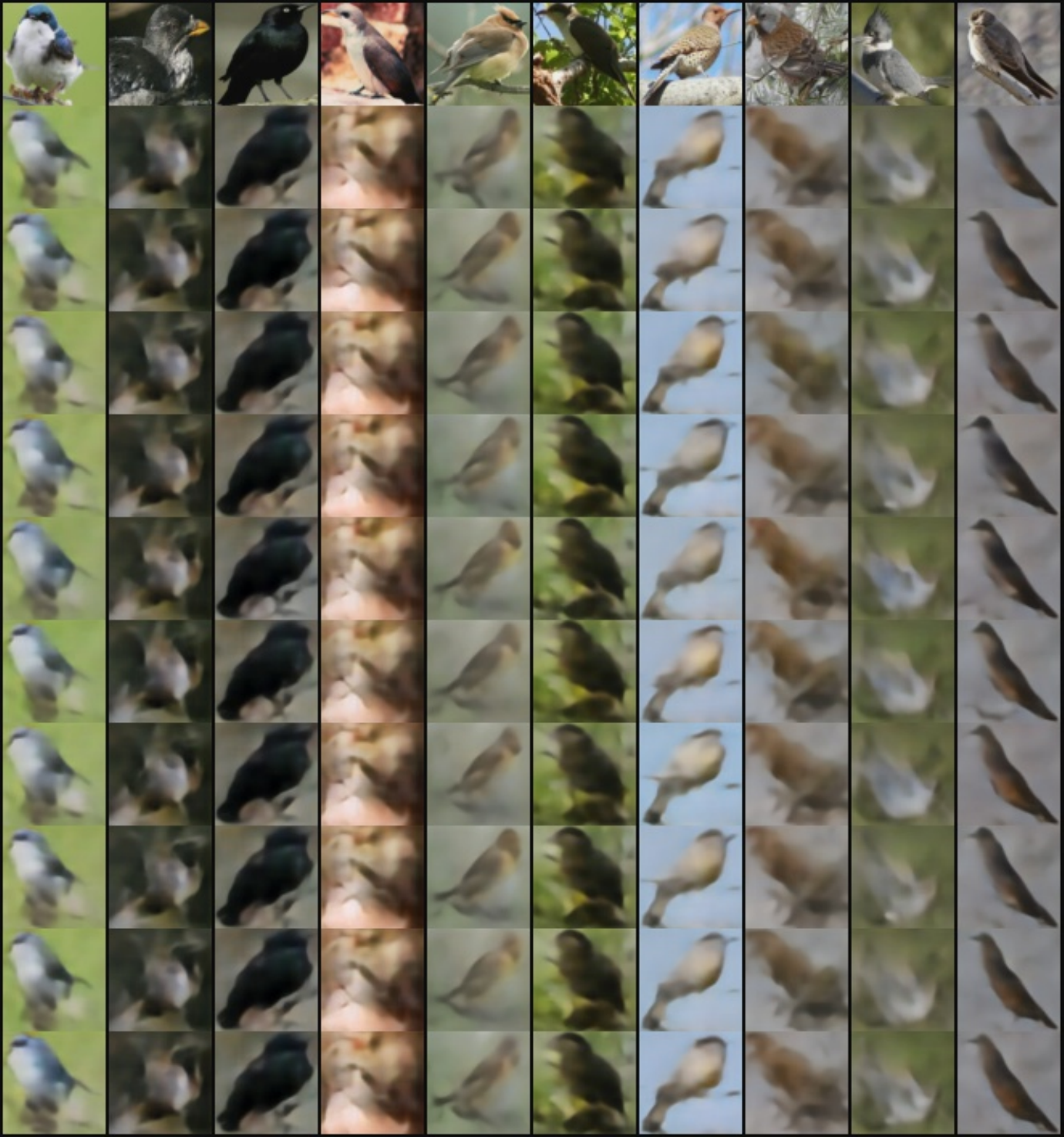}
    \caption{Visualization example 1: Unconditional self-reconstruction using the MVAE encoder and decoder.}
    \label{fig:vis1}
\end{figure}

\begin{figure}[t]
    \centering
    \includegraphics[width=0.45\textwidth]{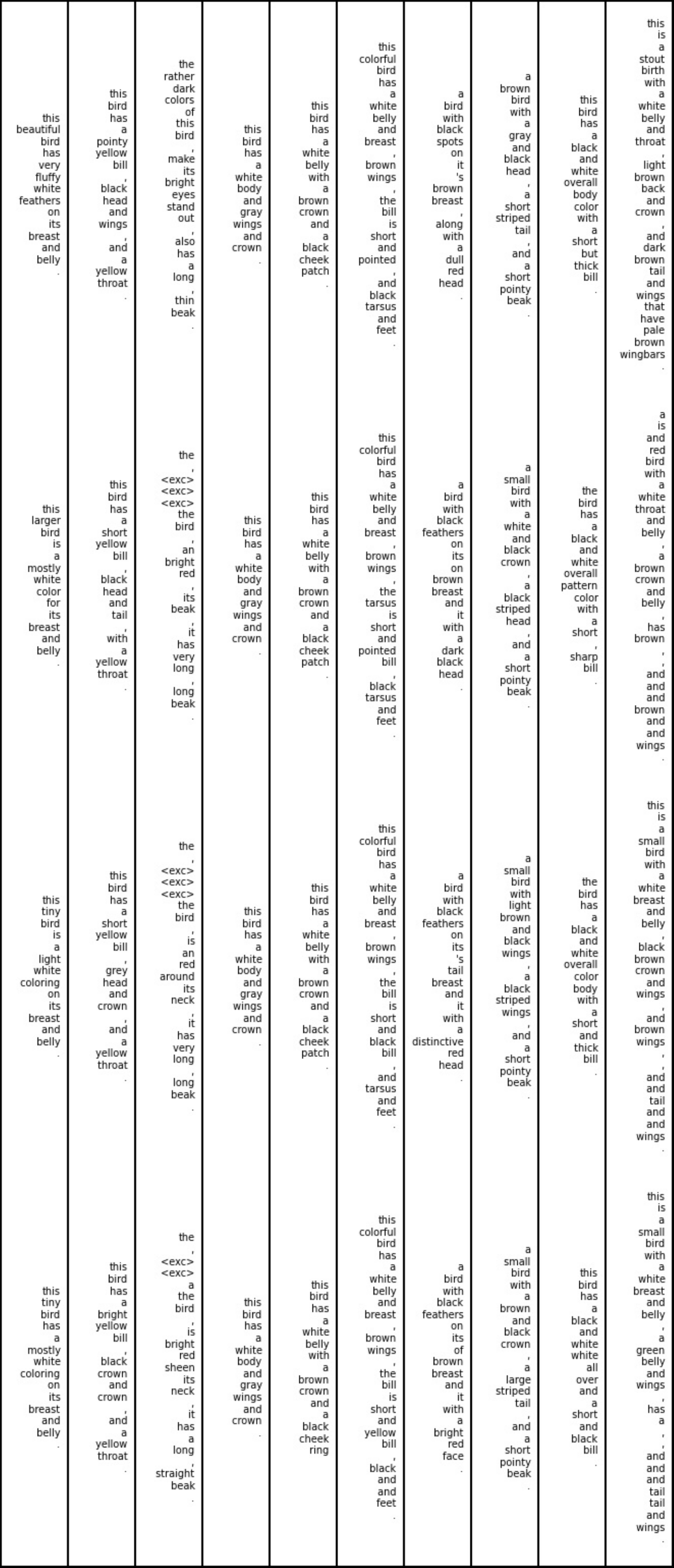}
    \caption{Visualization example 2: Unconditional self-reconstruction using the MVAE encoder and decoder.}
    \label{fig:vis2}
\end{figure}

\begin{figure}[t]
    \centering
    \includegraphics[width=0.45\textwidth]{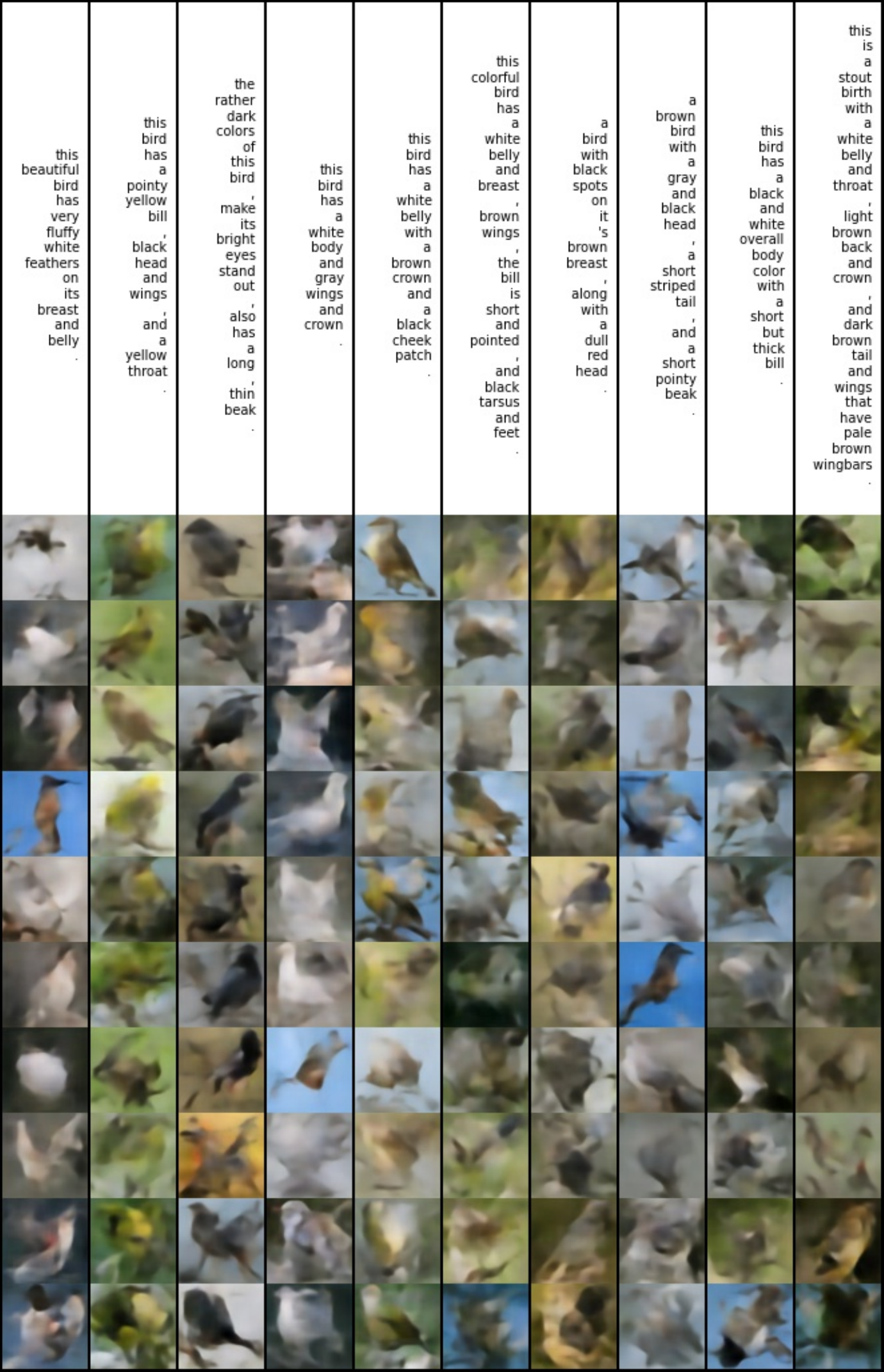}
    \caption{Visualization example 3: Cross-modal generation using our trained mapping network and MVAE backbone.}
    \label{fig:vis3}
\end{figure}

\begin{figure}[t]
    \centering
    \includegraphics[width=0.45\textwidth]{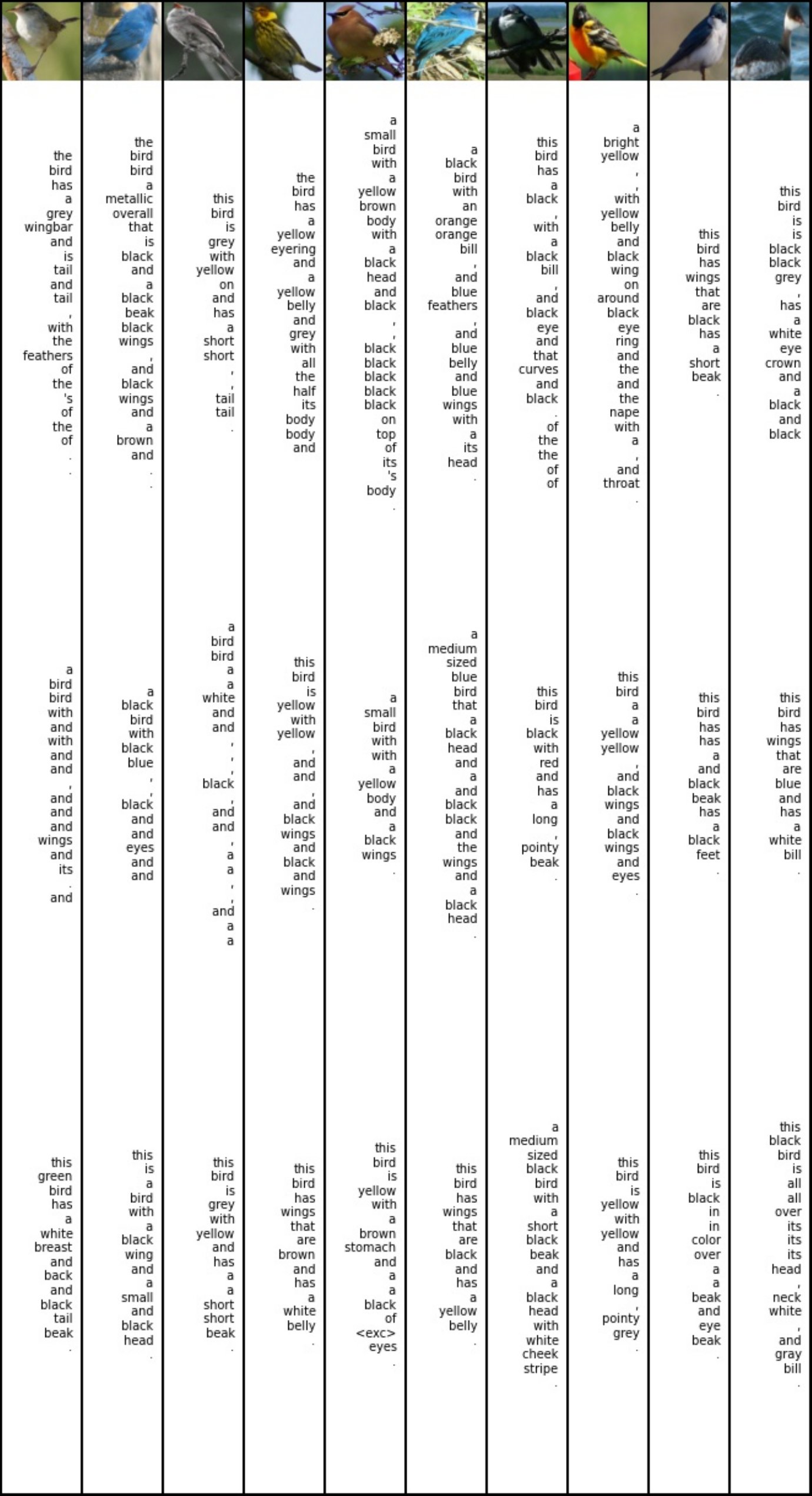}
    \caption{Visualization example 4: Cross-modal generation using our trained mapping network and MVAE backbone.}
    \label{fig:vis4}
\end{figure}

\begin{figure}[t]
    \centering
    \includegraphics[width=0.45\textwidth]{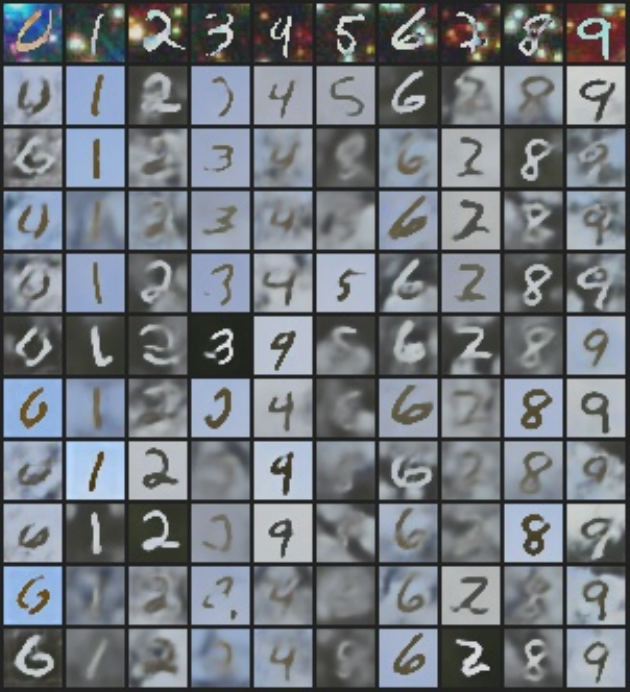}
    \caption{Visualization example 5: Cross-modal generation using our trained mapping network and MVAE backbone.}
    \label{fig:vis5}
\end{figure}

\begin{figure}[t]
    \centering
    \includegraphics[width=0.45\textwidth]{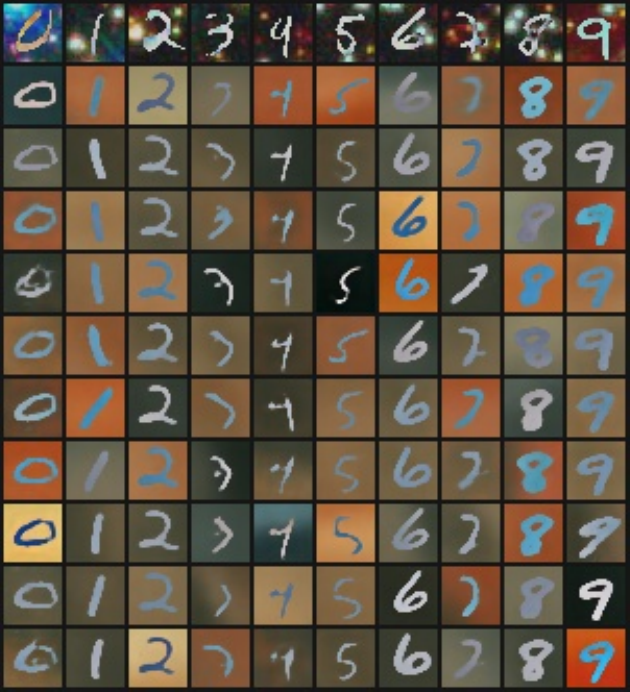}
    \caption{Visualization example 6: Cross-modal generation using our trained mapping network and MVAE backbone.}
    \label{fig:vis6}
\end{figure}

\begin{figure}[t]
    \centering
    \includegraphics[width=0.45\textwidth]{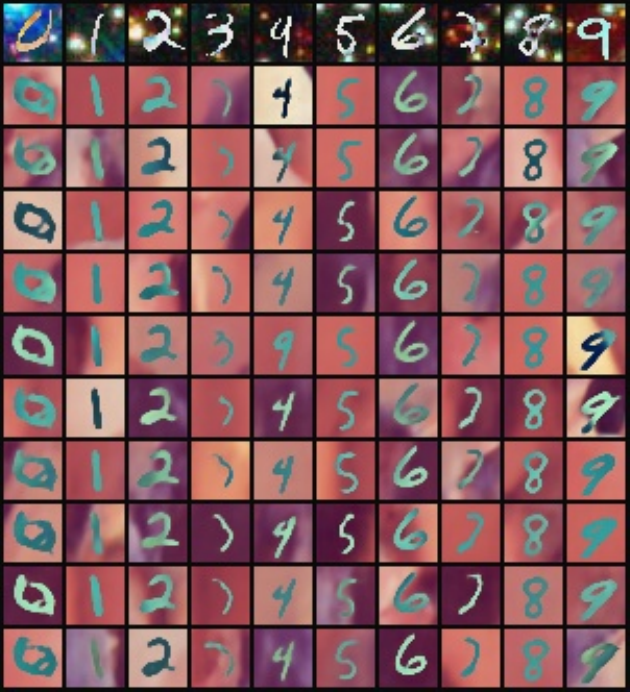}
    \caption{Visualization example 7: Cross-modal generation using our trained mapping network and MVAE backbone.}
    \label{fig:vis7}
\end{figure}

\begin{figure}[t]
    \centering
    \includegraphics[width=0.45\textwidth]{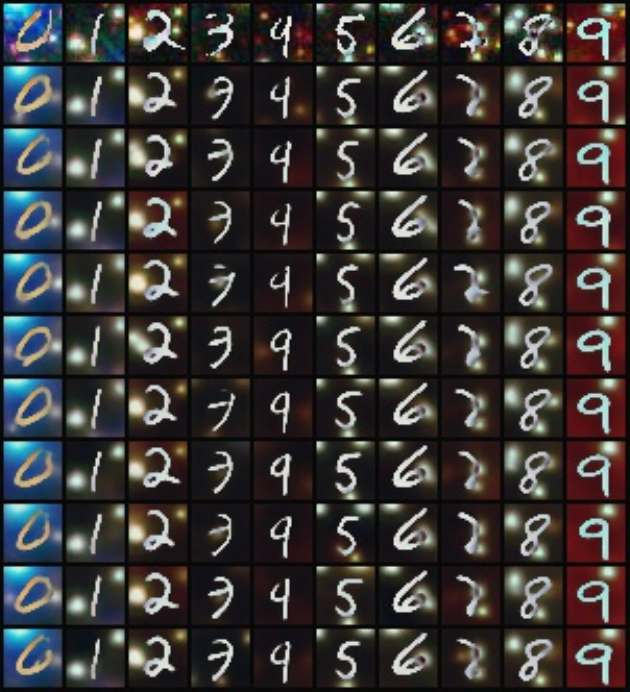}
    \caption{Visualization example 8: Cross-modal generation using our trained mapping network and MVAE backbone.}
    \label{fig:vis8}
\end{figure}

\begin{figure}[t]
    \centering
    \includegraphics[width=0.45\textwidth]{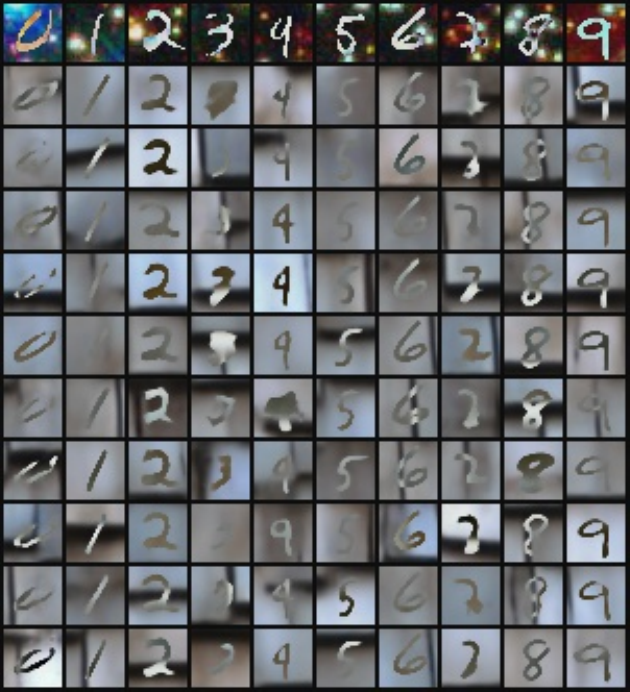}
    \caption{Visualization example 9: Cross-modal generation using our trained mapping network and MVAE backbone.}
    \label{fig:vis9}
\end{figure}

In this subsection, we provide additional qualitative results to further illustrate the reconstruction ability and cross-modal alignment of our framework. These visualizations are aimed at demonstrating how our method recovers missing modalities under diverse scenarios and dataset conditions.
Figures~\ref{fig:vis1}--\ref{fig:vis9} show generated results on the CUB and PolyMNIST datasets, where each sample illustrates how the trained modality-mapping network aligns latent representations across modalities.

\begin{figure}[t]
    \centering
    \includegraphics[width=0.48\textwidth]{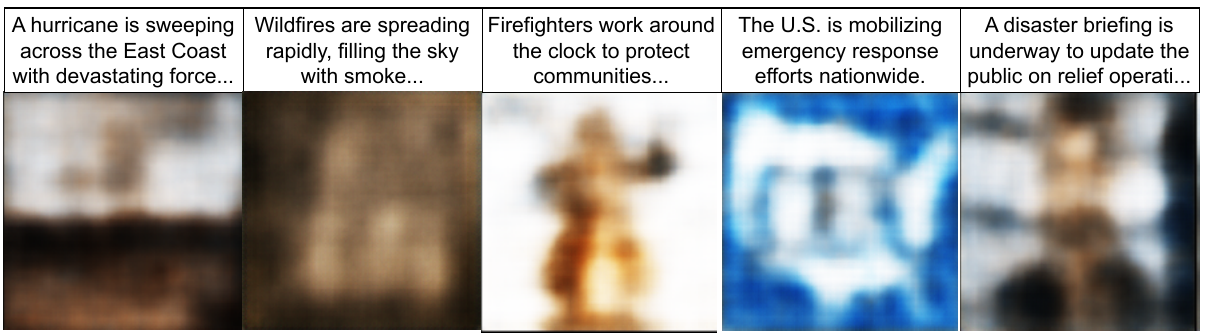}
    \caption{Examples of Image generated from Text on the CrisisMMD dataset, using the backbone trained with the MVAE framework.}
    \label{fig:CrisisMMD_TI}
\end{figure}

\begin{figure}[t]
    \centering
    \includegraphics[width=0.48\textwidth]{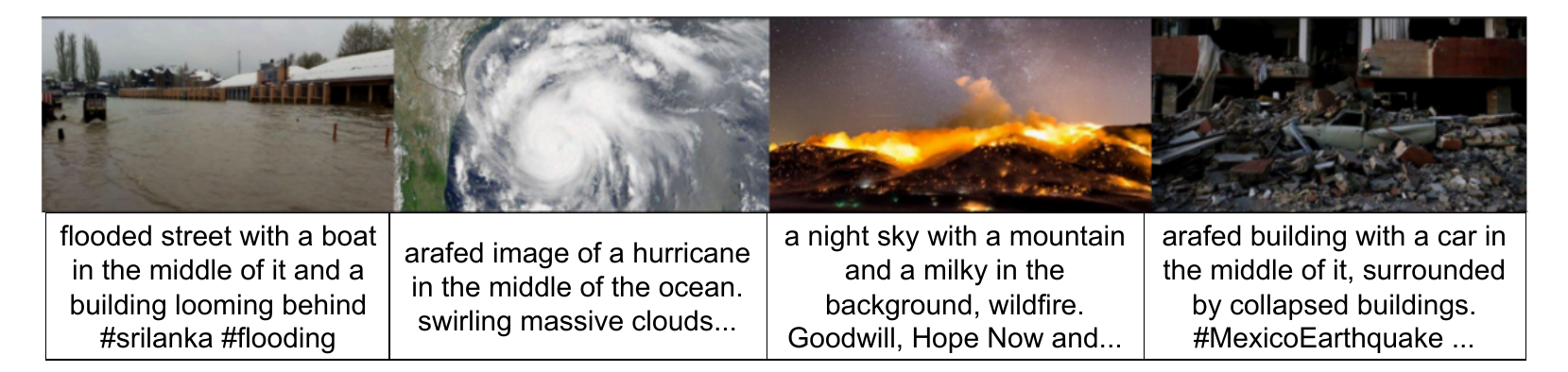}
    \caption{Examples of Text generated from Image on the CrisisMMD dataset, using the backbone trained with the MVAE framework.}
    \label{fig:CrisisMMD_IT}
\end{figure}

In addition, Figures~\ref{fig:CrisisMMD_TI} and \ref{fig:CrisisMMD_IT} present cross-modal generation results on the CrisisMMD dataset. Specifically, Figure~\ref{fig:CrisisMMD_TI} shows text-to-image generation, where the model generates visual content conditioned on input text. Figure~\ref{fig:CrisisMMD_IT} shows image-to-text generation, where the model synthesizes textual descriptions based on visual input. These examples demonstrate the model’s ability to perform bidirectional modality completion across vision and language, capturing both modality-specific detail and cross-modal semantic consistency.

\section{More Ablation Results}

To better understand the contribution of each component in our framework, we provide additional ablation results on the MELD and CrisisMMD datasets. Table~\ref{tab:ablation_meld} reports UAR scores on MELD under different proportions of missing modalities, while Table~\ref{tab:ablation_crisis} shows F1-scores on CrisisMMD when $\alpha=0.1$. In both cases, removing either the MMR module or the GGFS module leads to noticeable performance drops, highlighting their complementary roles. Notably, MMR contributes more significantly to the performance gain compared to GGFS.

\begin{table}[htbp]
\centering
\caption{UAR scores on MELD with different missing ratios. M: the cross-modal alignment reconstruction generator (MMR), G: the Global Generator Freezing Strategy (GGFS).}
\label{tab:ablation_meld}
\scalebox{0.65}{
\begin{tabular}{lccccccccc}
\toprule
\textbf{Setting} & \textbf{10\%} & \textbf{20\%} & \textbf{30\%} & \textbf{40\%} & \textbf{50\%} & \textbf{60\%} & \textbf{70\%} & \textbf{80\%} & \textbf{90\%} \\
\midrule
w/o M + G & 67.09 & 67.05 & 66.78 & 66.02 & 63.46 & 64.66 & 64.81 & 62.51 & 61.39 \\
w/o G      & 69.82 & 68.85 & 68.62 & 68.75 & 66.84 & 66.40 & 64.97 & 63.56 & 64.27 \\
FedRecon      & 70.11 & 69.94 & 69.65 & 69.18 & 67.12 & 66.53 & 65.28 & 63.91 & 64.81 \\
\bottomrule
\end{tabular}
}
\end{table}

\begin{table}[htbp]
\centering
\caption{F1 scores on CrisisMMD ($\alpha = 0.1$) under different missing ratios.}
\label{tab:ablation_crisis}
\scalebox{0.65}{
\begin{tabular}{lccccccccc}
\toprule
\textbf{Setting} & \textbf{10\%} & \textbf{20\%} & \textbf{30\%} & \textbf{40\%} & \textbf{50\%} & \textbf{60\%} & \textbf{70\%} & \textbf{80\%} & \textbf{90\%} \\
\midrule
w/o M + G & 37.53 & 38.29 & 38.68 & 36.06 & 35.51 & 32.13 & 30.33 & 28.08 & 15.36 \\
w/o G      & 41.06 & 41.97 & 41.52 & 39.85 & 38.01 & 34.82 & 32.55 & 28.37 & 18.94 \\
FedRecon      & 44.50 & 44.48 & 42.83 & 40.23 & 38.83 & 35.25 & 35.00 & 28.82 & 21.10 \\
\bottomrule
\end{tabular}
}
\end{table}

\section{GGFS Analysis and Visualization}

\begin{figure}[t]
    \centering
    \begin{subfigure}{0.23\textwidth}
        \centering
        \includegraphics[width=\textwidth]{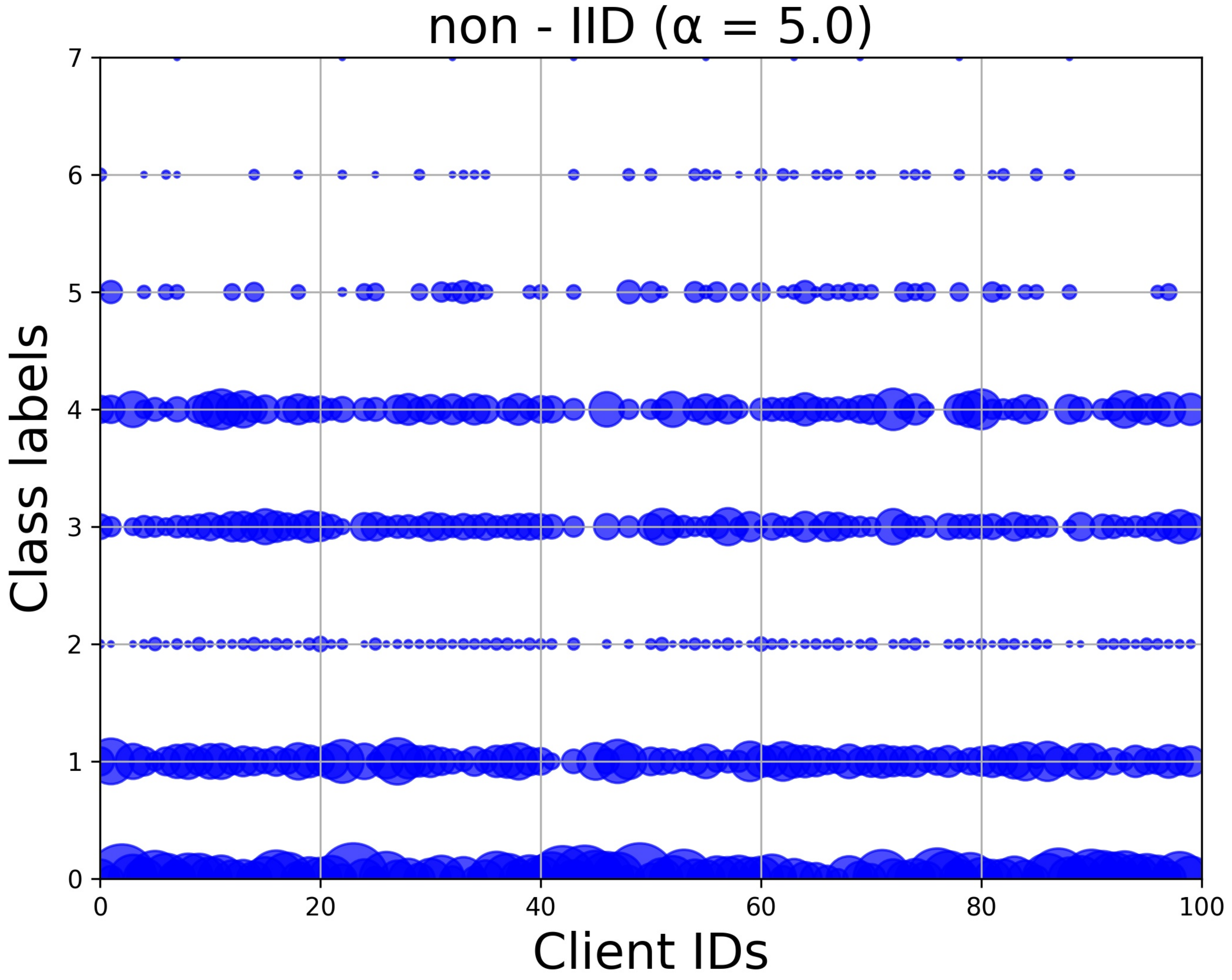}
        \caption{}
    \end{subfigure}
    \begin{subfigure}{0.23\textwidth}
        \centering
        \includegraphics[width=\textwidth]{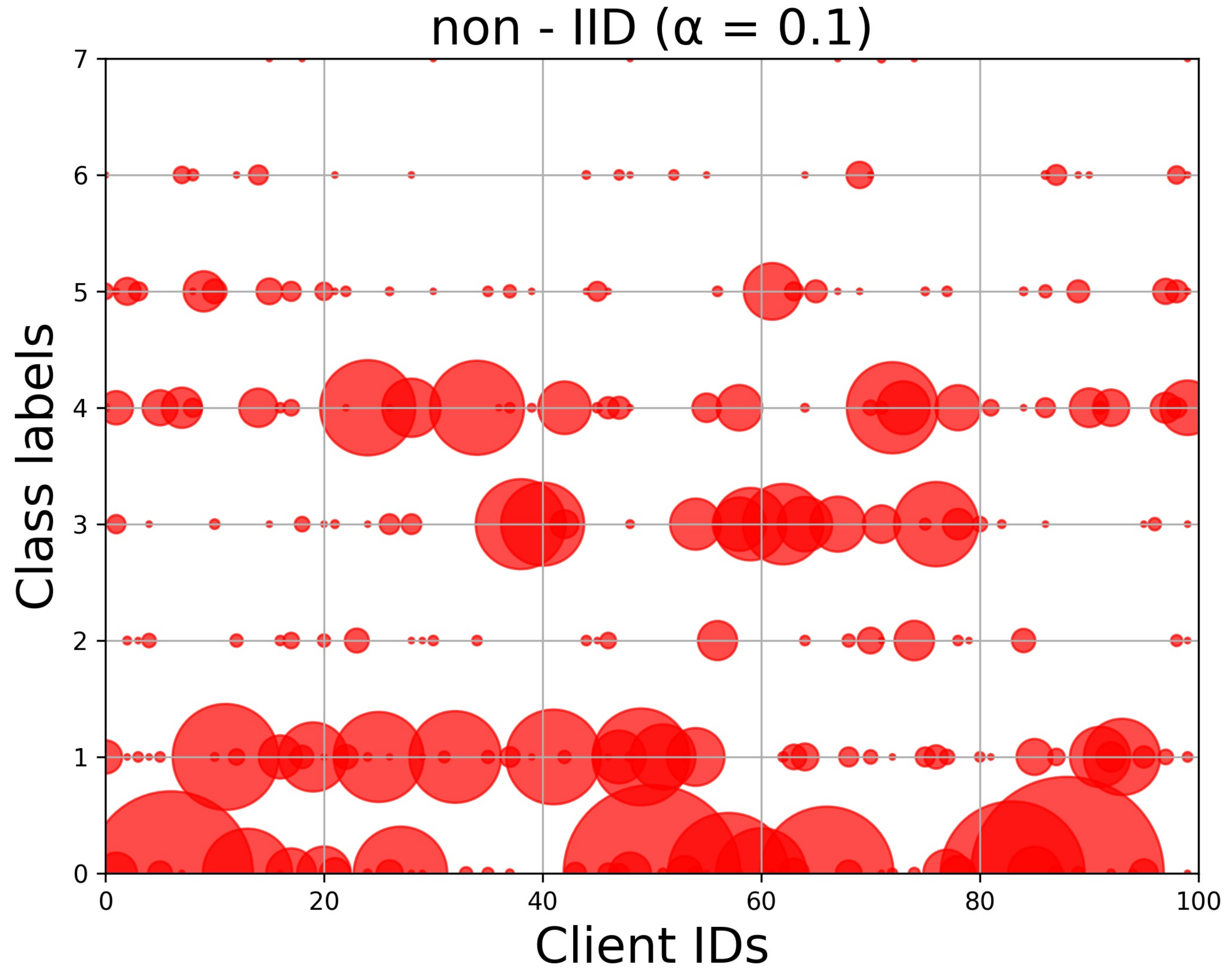}
        \caption{}
    \end{subfigure}
    \caption{Number of samples per class allocated to each client in CrisisMMD with $\alpha=5.0$ (a) and $\alpha=0.1$ (b).}
    \label{fig:label_distribution}
\end{figure}

We present the distribution of the number of samples per client in Figure~\ref{fig:label_distribution}, illustrating the scenario hypothesized in the main text. Specifically, some clients hold only a few samples that are difficult to reconstruct using their local generator, due to the presence of many other classes locally, which weakens the local generator’s ability to approximate these hard samples. In contrast, the global generator can better simulate these hard samples, as other clients possess a large number of similar hard samples.

\begin{figure}[t]
    \centering
    \includegraphics[width=0.48\textwidth]{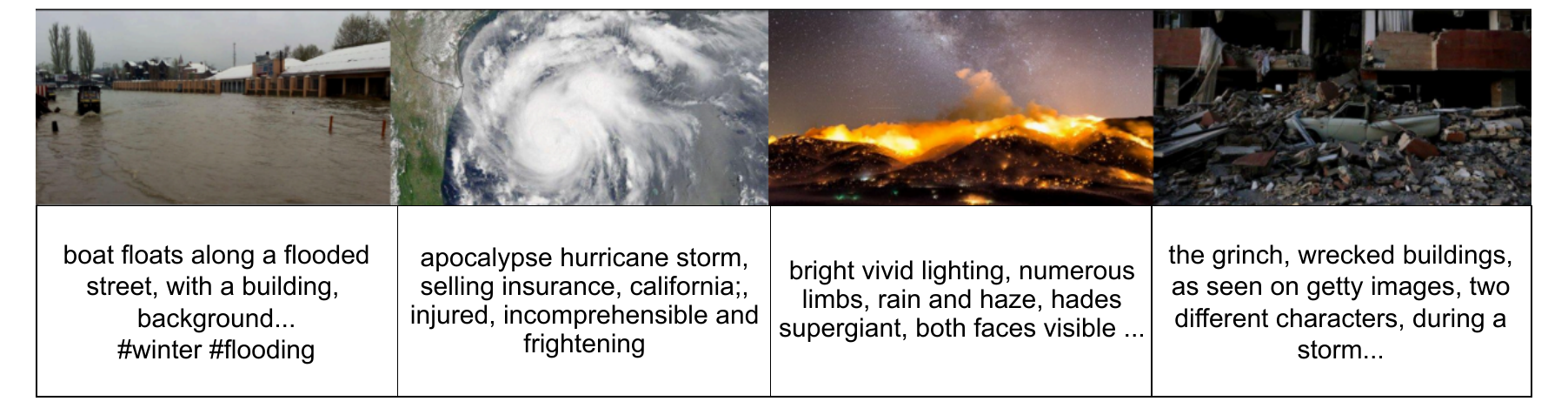}
    \caption{Examples of Text generated from Image on the CrisisMMD dataset, using the backbone trained with the MVAE framework without GGFS.}
    \label{fig:No_GGFS}
\end{figure}

Figure~\ref{fig:CrisisMMD_IT} shows results using GGFS, while Figure~\ref{fig:No_GGFS} presents results without GGFS. It can be observed that in the volcanic eruption scenario, the generator without GGFS produces misunderstandings.

\begin{table}[t]
\centering
\caption{Reconstruction of a missing modality on the PolyMNIST dataset using different combinations of available modalities.}
\label{tab:poly_reconstruction}
\begin{tabular}{lcc}
\toprule
\textbf{Available Modalities} & \textbf{Coherence ↑} & \textbf{FID ↓} \\
\midrule
$m_0$ & 0.839 & 76.87 \\
$m_1$ & 0.817 & 78.18 \\
$m_2$ & 0.834 & 79.20 \\
$m_3$ & 0.843 & 78.56 \\
$m_4$ & 0.845 & 78.92 \\
$m_0 + m_1$ & 0.847 & 79.29 \\
$m_3 + m_4$ & 0.858 & 78.18 \\
$m_1 + m_3 + m_4$ & 0.852 & 79.86 \\
$m_0 + m_1 + m_2 + m_3 + m_4$ & \textbf{0.861} & \textbf{80.21} \\
\bottomrule
\end{tabular}
\end{table}

We further examine whether a single available modality or multiple concatenated modalities should be used to reconstruct a missing one. As shown in Table~\ref{tab:poly_reconstruction}, on the PolyMNIST dataset, reconstruction using any single modality ($m_0$ to $m_4$) yields comparable performance. While incorporating multiple modalities can slightly improve both coherence and FID, the gain is marginal. Thus, using a single modality is generally sufficient and more efficient in terms of reconstruction cost.

\section{Training Cost Analysis}

We report the training cost of FedRecon across different datasets, as summarized in Table~\ref{tab:training_cost}. The total runtime consists of three components: MVAE, the mapping model, and the task model. Results show that the overall training time remains practical even on large-scale or high-resolution datasets.

\begin{table*}[t]
\centering
\caption{FedRecon runtime cost. Total training time is broken down into MVAE, Mapping Model, and Task Model components.}
\label{tab:training_cost}
\begin{tabular}{lcccc}
\toprule
\textbf{Model} & \textbf{Runtime/Round} & \textbf{Runtime/Round} & \textbf{Total Time} & \textbf{Total Time} \\
 & \textbf{(CrisisMMD)} & \textbf{(MELD)} & \textbf{(PolyMNIST)} & \textbf{(CUB)} \\
\midrule
MVAE & 26 s & 21 s & 189 min & 301 min \\
MMVAE & 33 s & 34 s & 237 min & 923 min \\
MMVAE+ & 35 s & 33 s & 249 min & 1036 min \\
Original Mapping Model & 4 s & 4 s & 37 min & 34 min \\
Simplified Mapping Model & 4 s & 4 s & \textbf{9 min} & 34 min \\
Task Model & 4 s & 3 s & 15 min & 17 min \\
\midrule
\textbf{Total (FedRecon)} & \textbf{34 s} & \textbf{28 s} & \textbf{241 min} & \textbf{352 min} \\
\bottomrule
\end{tabular}
\end{table*}

We primarily adopt the original mapping model in all experiments for consistency. However, we also evaluate a simplified variant that removes the dependency on source modality identity. As shown on PolyMNIST, this simplified version can further reduce training time with only marginal performance differences (see Appendix J).
In practice, we observe that both the MVAE and the mapping model tend to converge early, making early stopping an effective strategy to save computational cost. Additionally, when using pre-extracted features under the FedMultimodal setup, the peak GPU memory usage stays below \textbf{1000MB} even on a \textbf{GTX 1060}, enabling fast and memory-efficient training.

\section{Simplified Mapping Model Evaluation}

We simplify the modality mapping model by removing the dependency on the source modality index. We evaluate this simplified design on PolyMNIST, with results shown in Table~\ref{tab:simple_mapping}.

\begin{table}[t]
\centering
\caption{Evaluation of simplified mapping design on the PolyMNIST dataset. $T_{n,m}$ denotes a modality-specific mapping model, while $T_n$ uses a shared mapping regardless of the source modality.}
\label{tab:simple_mapping}
\begin{tabular}{lcc}
\toprule
\textbf{Model} & \textbf{Coherence ↑} & \textbf{FID ↓} \\
\midrule
$T_{n,m}$ & \textbf{0.845} & \textbf{78.92} \\
$T_n$     & 0.827          & 80.51 \\
\bottomrule
\end{tabular}
\end{table}

\section{Extension to MMVAE+ Backbone}

\begin{figure}[t]
    \centering
    \includegraphics[width=0.48\textwidth]{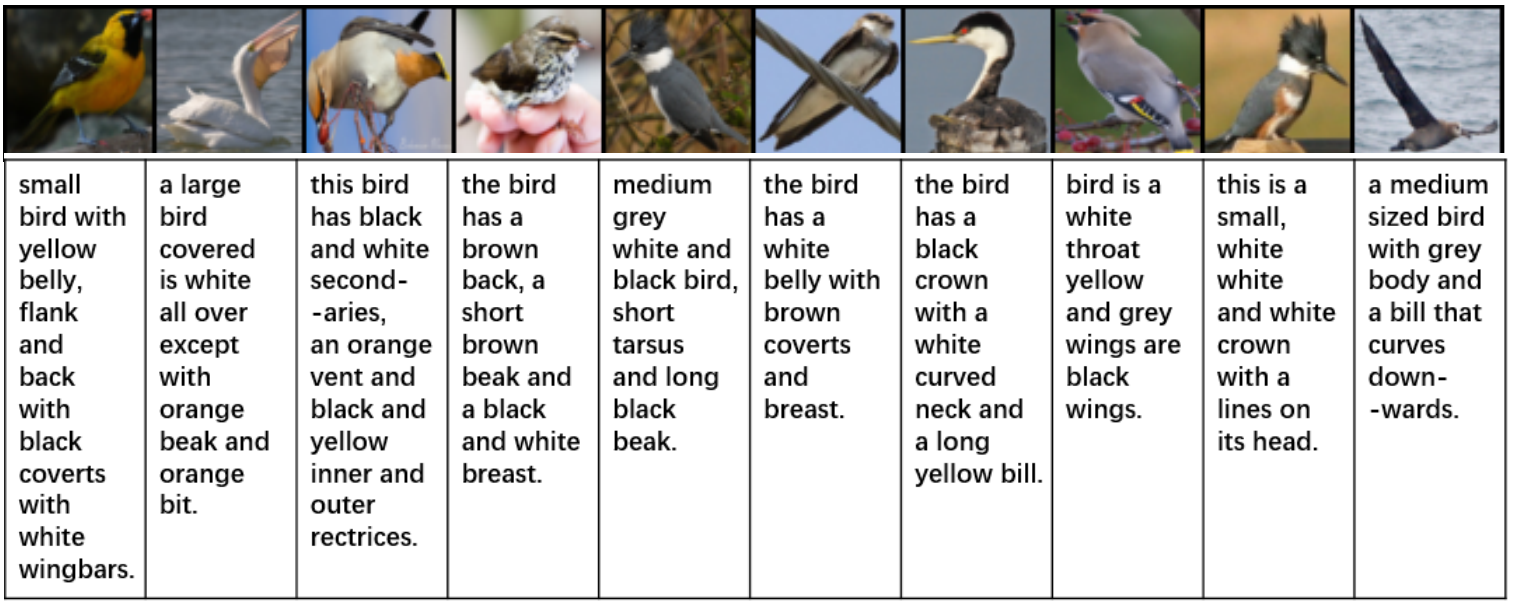}
    \caption{Visualization example 10: Cross-modal generation using our trained mapping network and MMVAE+ backbone.}
    \label{fig:vis10}
\end{figure}

We further evaluate FedRecon using MMVAE+ as the backbone encoder, demonstrating its compatibility with more expressive multimodal models. As shown in Figure~\ref{fig:vis10}, this setup yields improved coherence and generation quality under various qualitative scenarios.

\section{Comparison with CLAP}

CLAP is a pioneering work that first leverages MVAE for modality reconstruction under missing-modality settings. However, its missingness pattern differs fundamentally from ours. Specifically, CLAP assumes a “hard” missingness setup, where all samples within a client lack the same modalities, and the remaining modalities are complete and perfectly aligned (e.g., client 1 always lacks modality $X$, while having full access to $Y$ and $Z$). In contrast, FedRecon is designed for a more realistic “soft” missingness scenario, where the presence or absence of each modality varies on a per-sample basis within each client, typically following a Bernoulli distribution.

This distinction is critical: sample-level missingness reflects practical settings such as sensor dropouts or incomplete annotations, which are common in decentralized real-world applications. As discussed in the Introduction, FedRecon is the first to apply MVAE in such soft missing-modality federated settings.

Moreover, CLAP does not provide empirical analysis on why MVAE is preferable to other generative alternatives, especially under distributed constraints. In contrast, we provide experimental evidence to support the use of MVAE as a backbone for modality completion. While our setup is not fully aligned with CLAP’s original assumptions, it still satisfies the client-level connectivity requirements. Nonetheless, due to the high number of clients and the reduced per-client data under high missingness, CLAP’s performance degrades significantly, highlighting its limited applicability in realistic federated conditions.

\section{Limitations}

\textbf{Degradation of MVAE.} We train our MVAE generator using complete modality data. However, as the modality missing rate increases, the number of complete modality samples decreases, leading to a decline in modality alignment capability. When the missing rate reaches 100\%, MVAE degenerates into several unimodal VAEs. In our experimental setup, due to the distributional issues in modality partitioning, reconstruction remains a challenge. Specifically, when a single federated learning client experiences 100\% modality missing, reconstruction becomes exceedingly difficult. This is where GGFS comes into play, effectively preventing full degeneration by ensuring stable generative performance even under severe modality loss.

\textbf{Training Stability.} Another limitation of our method lies in the difficulty of achieving stable convergence for the mapping model. Interestingly, we observed conceptual similarities between our design and the training paradigm adopted in LLaVA \cite{Liu_2023_LLaVa}, which inspired us to adapt their training methodology. This proved beneficial—particularly for challenging datasets such as CrisisMMD, where we incorporated KL regularization to align the distributions of the two latent variables. This approach partially alleviated convergence issues.

We also experimented with a more direct solution by explicitly matching the means and variances of the latent distributions. While this method succeeded in narrowing the gap in statistical moments, the alignment was still imperfect and did not fully resolve the discrepancy in latent space structure. Notably, we observed that variance alignment was relatively easier to achieve, whereas the means tended to be less sensitive and exhibited smaller shifts.
We acknowledge these training instabilities as an area for improvement and plan to refine our optimization strategy in future work, with the aim of enhancing the robustness and reliability of \textbf{FedRecon}.

\section{Tricks}

We provide a potentially effective training trick: differentiating the latent distributions of different modalities. For instance, one could assign a Gaussian prior to modality A and a Laplacian prior to modality B. Alternatively, even if both modalities follow Gaussian distributions, their means and variances can be deliberately set to differ. As noted in\footnote{\url{https://spaces.ac.cn/archives/7725}}, the standard normal distribution is often the optimal default choice in the absence of distributional knowledge, due to its trainability and capacity to approximate diverse data distributions. However, if prior knowledge of the modality-specific distribution is available, we hypothesize that imposing distinct priors across modalities can enhance generative diversity and modality differentiation, ultimately improving generative coherence.

We conducted preliminary experiments across all four datasets using this trick, but the results exhibited high variance. We believe that with appropriately curated datasets or more stable training settings, this approach holds promise and could be practically beneficial.

\end{document}